\documentclass{article}

\PassOptionsToPackage{numbers, compress}{natbib}
\usepackage[preprint]{neurips_2026}
\usepackage[utf8]{inputenc}
\usepackage[T1]{fontenc}
\usepackage{hyperref}
\usepackage{url}
\usepackage{booktabs}
\usepackage{longtable}
\usepackage{amsfonts}
\usepackage{nicefrac}
\usepackage{microtype}
\usepackage{xcolor}
\usepackage{colortbl}
\usepackage{amsmath,amssymb}
\usepackage{graphicx}
\graphicspath{{figures/}}
\usepackage{wrapfig}
\usepackage{algorithm}
\usepackage{algorithmic}
\usepackage{placeins}

\newcommand{\methodname}{Taylor-Calibrate}
\newcommand{\widetablefont}{\fontsize{6.0}{7.2}\selectfont}

\title{\methodname: Principled Initialization for Hybrid Linear Attention Distillation}

\author{
Zhongzhu Zhou\textsuperscript{1,2}
\quad Qingyang Wu\textsuperscript{2}
\quad Junxiong Wang\textsuperscript{5}
\quad Mayank Mishra\textsuperscript{3} \\
Shuaiwen Leon Song\textsuperscript{2}
\quad Ben Athiwaratkun\textsuperscript{2}
\quad Chenfeng Xu\textsuperscript{2,4} \\
\textsuperscript{1}The University of Sydney
\quad \textsuperscript{2}Together AI \\
\textsuperscript{3}University of California, Berkeley
\quad \textsuperscript{4}The University of Texas at Austin \\
\textsuperscript{5}Microsoft
}

\begin{document}
\maketitle

\begin{abstract}
Hybrid linear attention models offer an appealing path to faster long-context inference: they reduce the quadratic cost and KV-cache burden of full softmax attention while retaining much of the quality of Transformer models. A practical way to obtain such models is to convert a pretrained Transformer instead of pretraining a new architecture from scratch, but this conversion is still brittle. Simply copying the teacher attention projections into a Gated DeltaNet (GDN) student does not specify the new recurrent decay, write, and output-gating dynamics. As a result, the converted model often starts in a poor dynamical regime and must spend many distillation tokens repairing initialization rather than learning the remaining teacher behavior. We propose \methodname{}, a lightweight initialization method for hybrid GDN students. The method uses Taylor-guided teacher attention statistics to set the value projection, memory timescale, write gates, and output gate, then applies a short per-layer alignment step to match each converted layer to the teacher output. Across four teacher settings and three retained-layer policies, \methodname{} gives substantially stronger zero-shot students, with up to an 88$\times$ improvement in a representative ablation, and reaches matched recovery targets with 4.9$\times$--9.2$\times$ fewer training tokens than naive conversion.\footnote{Code is available at \url{https://github.com/FutureMLS-Lab/Taylor-Calibrate}.}
\end{abstract}

\section{Introduction}
\label{sec:intro}

Transformers remain the default choice for large language models, but long-context decoding exposes a practical weakness: attention is quadratic, and the KV cache grows with sequence length \cite{vaswani2017attention}. This has pushed the community toward efficient sequence mixers that keep a fixed-size recurrent or state-space memory. Early kernelized and fast-weight views already showed that attention-like aggregation can be written as a recurrent computation \cite{katharopoulos2020transformers,schlag2021linear,peng2022abc}; more recent work builds on this idea with selective state spaces and gated linear attention \cite{gu2024mamba,dao2024ssm,yang2024gla,zhang2024hedgehog,liu2024longhorn}. Large open models such as Jamba, Jamba-1.5, Samba, Hymba, B'Mojo, MiniMax-01, Falcon-H1, Kimi Linear, and Qwen3.5 further suggest that replacing many softmax layers can improve the trade-off among quality, throughput, KV-cache cost, and long-context support \cite{lieber2024jamba,jamba2024jamba15,ren2024samba,dong2024hymba,zancato2024bmojo,minimax2025minimax01,falcon2025h1,kimi2025linear,qwen2026qwen35}, which makes hybridization attractive. Still, no single efficient mixer works best everywhere: pure recurrent or linear models can struggle on retrieval-heavy and associative-recall tasks, while hybrid models can preserve some softmax layers and use cheaper recurrent modules elsewhere \cite{rao2024justreadtwice,wen2025rnns,irie2025blending,wang2025hybridanalysis,yang2025zebrallama}.

Rather than training a hybrid model from scratch, recent work often starts from a pretrained Transformer and converts part of it into a linear or recurrent student. Some methods swap the attention block then continue training \cite{kasai2021finetuning,mercat2024linearizing,lan2025liger,nguyen2025lizard}; others add attention transfer, hidden-state matching, or staged distillation before full recovery training \cite{wang2024mamba,zhang2025lolcats,goldstein2025radlads,bick2025llamba,bick2026retrievalaware}. Hybrid-specific work keeps only a subset of layers as softmax attention before targeted distillation \cite{hoshino2025rad,li2025klguided}. These pipelines reuse useful teacher structure, but the new recurrent dynamics are often left for later training to discover. In practice, this makes conversion sensitive to the initialization, token budget, and training schedule. Appendix~\ref{sec:related} gives a more detailed comparison to prior conversion and layer-selection methods, including RADLADS~\cite{goldstein2025radlads} and GA-S2~\cite{li2025klguided}.

The central issue is that hybrid transfer is not only a projection-copying problem. Copying the teacher's $Q/K/V/O$ weights~\cite{wang2024mamba} preserves useful projections, but it does not specify the recurrent timescales, write gates, or output gates that now control information flow. This matters especially for Gated DeltaNet (GDN) \cite{yang2024gdn}, where the recurrent output is also modulated by a learned output gate before the final projection. If these new parameters are left random, the converted student can have very poor zero-shot quality even when the inherited projections are reasonable. Linear and recurrent mixers can be viewed as fixed-memory approximations to the same causal aggregation computed by softmax attention. A low-order Taylor view makes this connection operational: value amplitude, average look-back distance, and attention concentration provide teacher-side proxies for the value scale, recurrent decay timescale, and write strength of the GDN state. This motivates an initialization objective beyond projection copying: use teacher attention statistics to set the decay, write, and output-gate scales before expensive full-model distillation begins.

We propose \textbf{\methodname}, a lightweight two-stage initialization method for GDN-based hybrid distillation. The first stage reads simple statistics from the teacher attention maps and turns them into GDN parameter settings: value scale initializes $W_V$, average attention distance sets how slowly the recurrent state should decay, and attention concentration sets the write gate. The second stage runs a short layer-local alignment step, using calibration data to make each converted GDN layer match the teacher layer output. This makes initialization an explicit transfer step, rather than leaving downstream distillation to discover the right recurrent dynamics from scratch. 

\begin{figure}[t]
    \centering
    \begin{minipage}[t]{0.49\linewidth}
    \centering
    \textbf{(a) Zero-shot initialization}\\[2pt]
    \includegraphics[width=\linewidth]{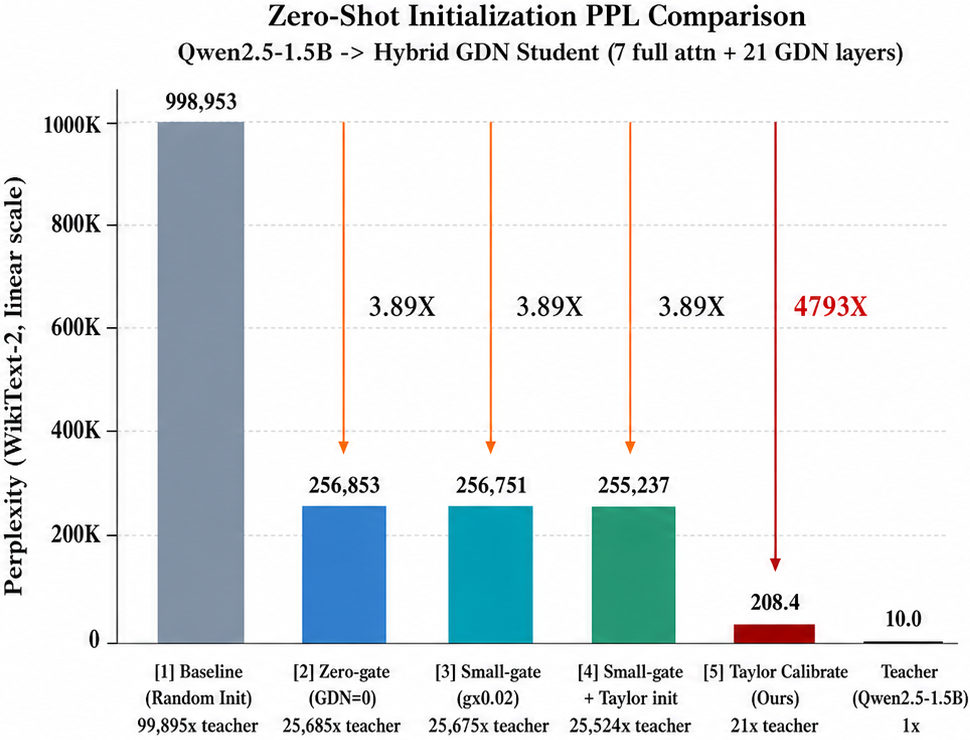}
    \end{minipage}%
    \hfill
    \begin{minipage}[t]{0.49\linewidth}
    \centering
    \textbf{(b) Qwen3-8B training loss}\\[2pt]
    \includegraphics[width=\linewidth]{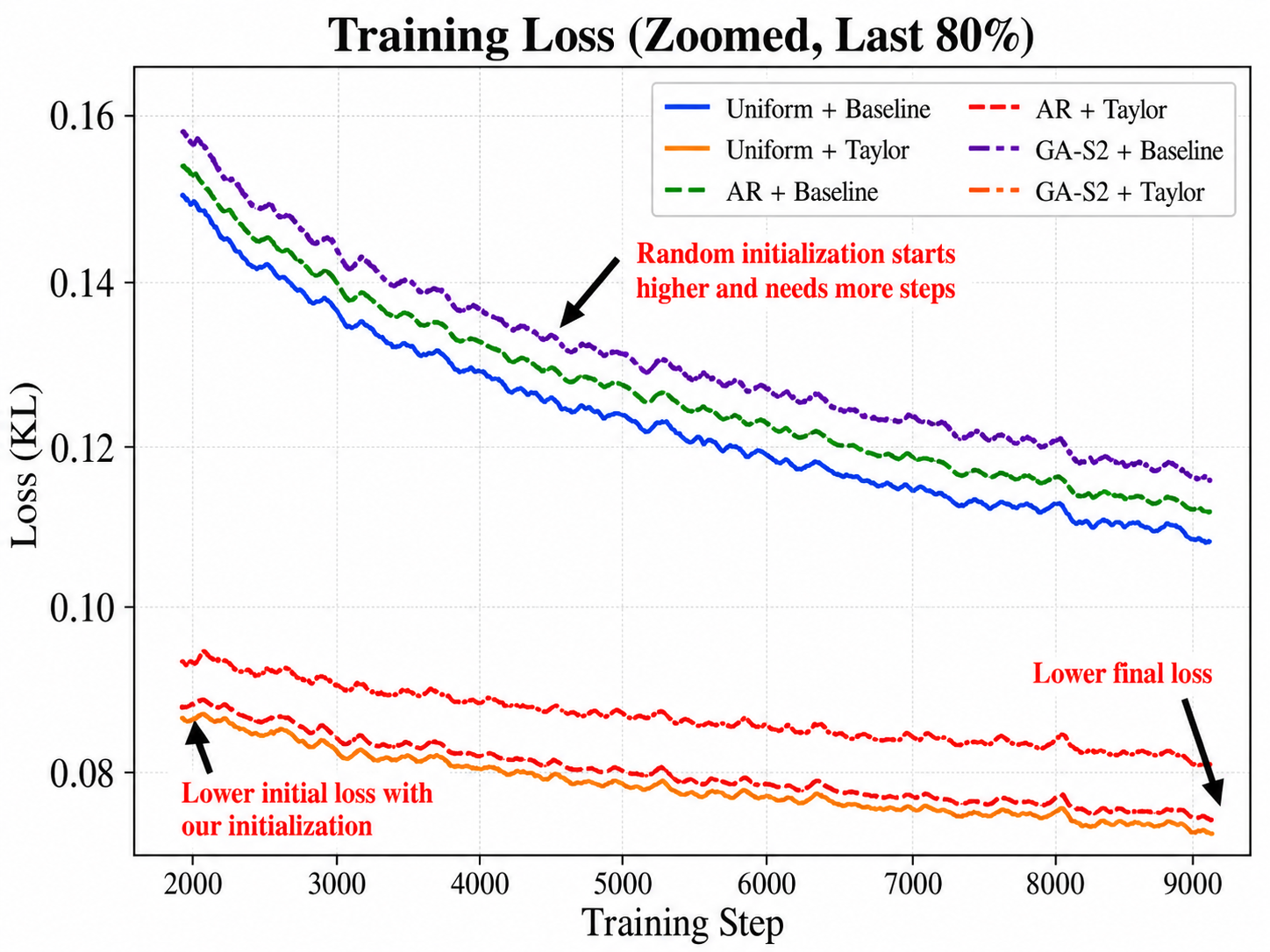}
    \end{minipage}
    \caption{Motivating examples for hybrid GDN initialization. \textbf{Left:} zero-shot PPL is highly sensitive to how the newly introduced recurrent parameters are initialized, even when all variants reuse the same teacher $Q/K/V/O$ projections. \textbf{Right:} in an actual Qwen3-8B training run, \methodname{} starts from a lower loss and maintains a better optimization trajectory than random initialization.}
    \label{fig:intro-motivation}
    \vspace{-1em}
    \end{figure}

Figure~\ref{fig:intro-motivation} shows the key motivation. In representative conversions, all variants inherit the same teacher $Q/K/V/O$ projections, but their zero-shot quality changes sharply with the GDN gate initialization. The training curves show the same pattern after distillation begins: better initialization starts from a lower loss and keeps a more favorable optimization trajectory, suggesting that the converted model is not merely under-trained but starts in the wrong dynamical regime.

Although we focus on GDN, the broader message is that linear-attention transfer needs to initialize \emph{dynamical parameters}, not just copy projection weights.

Our main contributions are:
\begin{itemize}
    \item We formulate hybrid linear-attention conversion as an initialization problem for recurrent dynamics, not just a projection-copying problem.
    \item We introduce \methodname{}, a two-stage GDN initialization method that combines Taylor-derived gate calibration with brief layer-local alignment.
    \item We evaluate across four teacher settings and three layer-selection policies, reporting higher zero-shot Avg and faster downstream recovery than naive initialization.
\end{itemize}
\section{Preliminaries}
\label{sec:prelim}

\subsection{Softmax Attention}

Consider a decoder-only Transformer \cite{vaswani2017attention} with hidden states $x_1,\ldots,x_T \in \mathbb{R}^d$. For one attention head with head dimension $d_h$, softmax attention computes
\begin{equation}
q_t = W_Q x_t,\qquad k_t = W_K x_t,\qquad v_t = W_V x_t,
\end{equation}
and outputs
\begin{equation}
o_t = \sum_{s \le t} \alpha_{t,s} v_s,\qquad
\alpha_{t,s} = \frac{\exp(q_t^\top k_s / \sqrt{d_h})}{\sum_{j \le t} \exp(q_t^\top k_j / \sqrt{d_h})}.
\end{equation}
During autoregressive decoding, the hidden states of all previous keys and values must be retained, leading to a KV cache that grows linearly with sequence length. This cache is a central systems bottleneck for long-context inference \cite{wang2024mamba,kimi2025linear}.

\subsection{Taylor View of Softmax Attention}

A useful theoretical lens for our method is to expand the softmax numerator around a neutral logit regime, following the broader view that softmax attention admits linear-attention and recurrent approximations \cite{katharopoulos2020transformers,schlag2021linear,peng2022abc}:
\begin{equation}
\exp(s) = 1 + s + \frac{s^2}{2} + \mathcal{O}(s^3).
\end{equation}
Truncating this expansion motivates a fixed-size linear-recurrent approximation of softmax attention. More importantly for our purposes, it yields a \emph{calibration methodology}: second-order curvature can be compiled into a small number of effective scaling terms. One convenient summary of this viewpoint is that the second-order correction can be folded into an effective first-order sequence-mixing scale, often written as $\gamma^2 = 1 + \mu_3/(2\mu_2)$ for low-order logit moments $\mu_2,\mu_3$, while the value pathway admits a per-head least-squares rescaling factor $\sigma$ that matches the teacher output amplitude.


\subsection{Gated DeltaNet (GDN) Linear Attention}

Gated DeltaNet replaces softmax attention with a recurrent state update \cite{yang2024gdn}. In the variant used here, queries and keys are projected from the same teacher-inherited hidden states and then normalized per head. Let $S_t \in \mathbb{R}^{d_h \times d_h}$ denote the recurrent memory state. For each step,
\begin{equation}
g_t = -\exp(A_{\log}) \, \operatorname{softplus}(a_{\mathrm{proj}}(x_t) + dt_{\mathrm{bias}}),
\qquad
\beta_t = \sigma(b_{\mathrm{proj}}(x_t)),
\end{equation}
\begin{equation}
S_t = \exp(g_t)\left(S_{t-1} - \beta_t S_{t-1} k_t k_t^\top\right) + \beta_t v_t k_t^\top,
\qquad
o_t = S_t q_t.
\end{equation}
The final output is further modulated by an output gate,
\begin{equation}
y_t = W_O\!\left(\operatorname{RMSNorm}(o_t) \odot \operatorname{SiLU}(g_{\mathrm{proj}}(x_t))\right).
\end{equation}
Relative to softmax attention, GDN removes the explicit KV cache and instead compresses context into a fixed-size recurrent state \cite{yang2024gdn}. This provides constant-memory inference with respect to context length, but places substantial burden on the decay and gating parameters. If these parameters are poorly initialized, the recurrent block may either erase useful history too aggressively or inject unstructured noise into the residual stream.

\section{Method: \methodname}
\label{sec:method}

\begin{figure}[t]
\centering
\includegraphics[width=\linewidth]{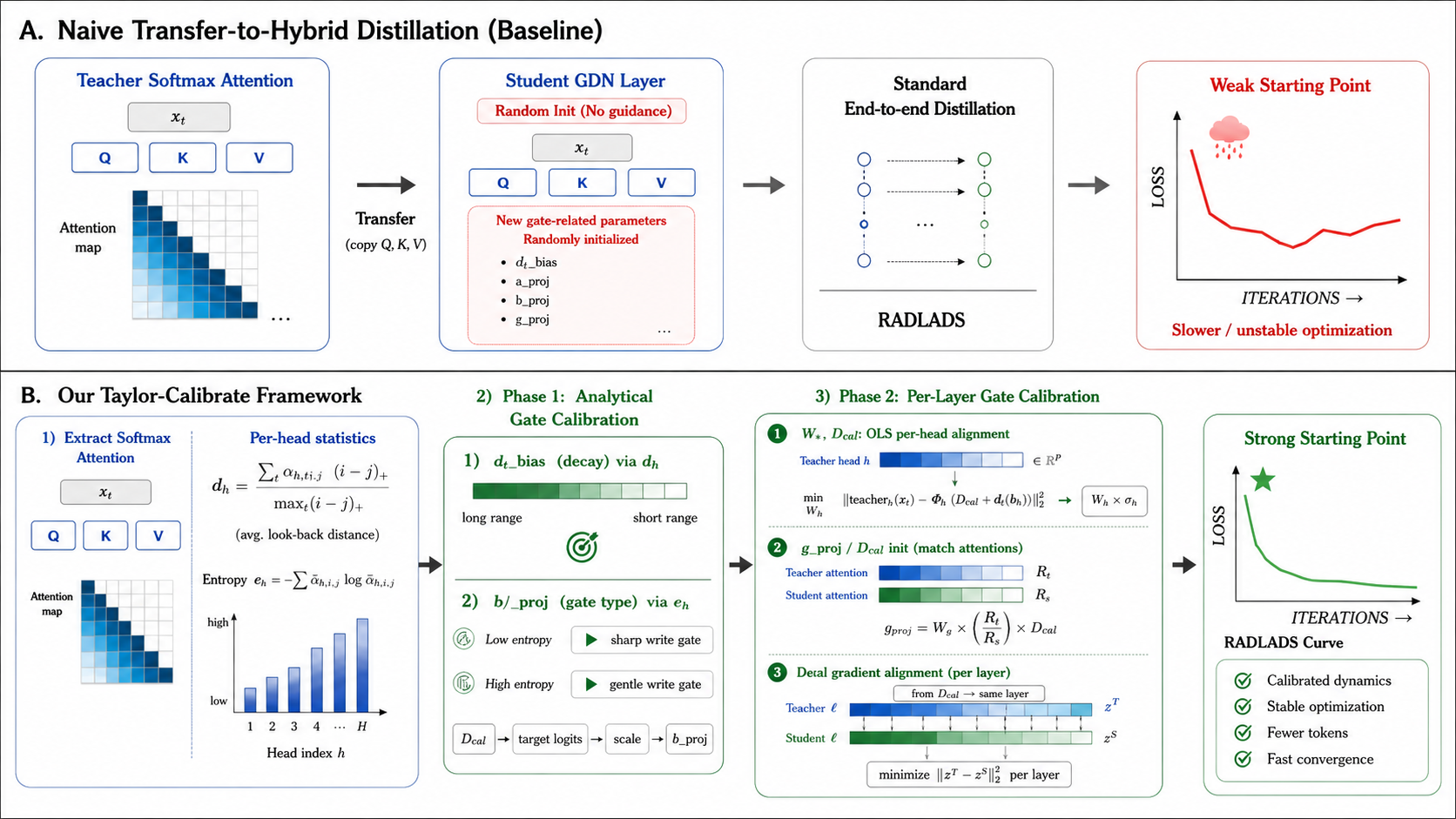}
\vspace{-0.4cm}
\caption{Overview of \methodname. Standard transfer copies teacher projections but leaves new GDN dynamics at default initialization. \methodname{} instead uses teacher attention statistics to calibrate recurrent gates, then performs a short per-layer alignment step before downstream distillation.}
\label{fig:method-overview}
\end{figure}

\methodname{} initializes a converted GDN layer \cite{yang2024gdn} in two steps. First, we use Taylor-guided statistics from the teacher attention map, motivated by linear-attention approximations to softmax attention \cite{katharopoulos2020transformers,schlag2021linear,peng2022abc}, to set the value scale, decay bias, write gate, and output gate. Second, on the same calibration inputs, we align the GDN layer output to the teacher layer output before global distillation, following the hidden-state matching used in recent conversion pipelines \cite{wang2024mamba,zhang2025lolcats,goldstein2025radlads}. Figure~\ref{fig:method-overview} gives the full pipeline, and Table~\ref{tab:gdn-init-params} lists the parameters affected by this transfer. Appendix~\ref{app:calibration} formalizes the hybrid setting and provides the derivations behind these calibration rules.

\begin{table}[t]
\centering
\footnotesize
\caption{Initialization targets for a converted GDN layer. Standard transfer copies inherited projections but leaves new recurrent dynamics at default values; \methodname{} first extracts Taylor-derived calibration statistics, then initializes GDN parameters before local alignment.}
\label{tab:gdn-init-params}
\begin{tabular}{@{}p{0.20\linewidth}p{0.32\linewidth}p{0.38\linewidth}@{}}
\toprule
Parameter/statistic & Standard GDN transfer & \methodname{} initialization \\
\midrule
$d_h,e_h,c_h,\sigma_h^\star$ & Not used. & Teacher attention statistics. \\
$W_Q,W_K,W_O$ & Copy from teacher. & Copy from teacher. \\
$W_V$ / value path & Copy from teacher. & $W_V \leftarrow \sigma_h^\star W_V$ (OLS scale). \\
$A_{\log}$ & Default decay scale. & Default decay scale. \\
$dt_{\mathrm{bias}}$ & Default time bias. & $dt_{\mathrm{bias},h}=\operatorname{softplus}^{-1}(\ln 2/d_h)$. \\
$a_{\mathrm{proj}}$ & Default decay projection. & Default decay projection. \\
$b_{\mathrm{proj}}$ & Default write projection. & Row scale to $\operatorname{logit}(\beta_h^\star)$, $\beta_h^\star=0.3+0.4c_h$. \\
$g_{\mathrm{proj}}$ & Default or small-gate heuristic. & Small/RMS-matched gate. \\
\bottomrule
\end{tabular}
\end{table}

\subsection{Problem Setup}

Let $M_T$ be a pretrained Transformer teacher \cite{vaswani2017attention} and $M_S$ a hybrid student that keeps some softmax layers and replaces the rest with GDN \cite{yang2024gdn}; Appendix~\ref{app:hybrid_architectures} states this layer partition explicitly. For each converted layer $\ell \in \mathcal{S}_{\mathrm{gdn}}$, we compare the teacher sequence-mixing output $z_T^{(\ell)}(x)$ with the student GDN output $z_S^{(\ell)}(x;\theta_\ell)$ on the same calibration input $x$. The initialization problem is to make this local mismatch small before any expensive full-model training:
\begin{equation}
\mathcal{L}_{\mathrm{init}} =
\sum_{\ell \in \mathcal{S}_{\mathrm{gdn}}}
\mathbb{E}_{x \sim \mathcal{D}_{\mathrm{cal}}}
\left\| z_S^{(\ell)}(x; \theta_\ell) - z_T^{(\ell)}(x) \right\|_2^2,
\end{equation}
where $\mathcal{D}_{\mathrm{cal}}$ is a small calibration corpus. We do not try to solve end-to-end distillation here; the objective is only to reduce the layer-local mismatch before full-model training.

This formulation is motivated by a direct observation from conversion experiments: even when students inherit the same teacher $W_Q,W_K,W_V,W_O$, their initial PPL and task accuracy can vary sharply depending on how the newly introduced GDN gates are set. The missing transfer information is therefore in the recurrent dynamics: memory timescale, write strength, and output gating. \methodname{} uses a Taylor view of the teacher attention map to choose proxies for these quantities, then uses a short layer-local alignment step for the remaining nonlinear mismatch.

\subsection{Phase 1: Taylor-Derived Calibration Adapted to GDN}

Phase 1 keeps only the Taylor quantities that map cleanly to GDN parameters. Teacher output amplitude calibrates the value path, average attention distance calibrates the decay bias, attention entropy calibrates the write gate through a concentration score, and the output gate is initialized small enough not to overwhelm the recurrent state. This adapts the recurrent view of attention \cite{katharopoulos2020transformers,schlag2021linear,peng2022abc} to the specific decay and gating structure of GDN \cite{yang2024gdn}.

Concretely, for each converted layer and attention head, we compute the average attention distance $d_h=\mathbb{E}[\sum_{s \le t} a_{t,s}^{(h)}(t-s)]$ and entropy $e_h=\mathbb{E}[-\sum_{s \le t} a_{t,s}^{(h)}\log a_{t,s}^{(h)}]$, where $a_{t,s}^{(h)}$ is the teacher attention probability from position $t$ back to position $s$ in head $h$. The distance statistic $d_h$ is the attention-weighted average look-back length: a large value means the teacher head often reads far into the past, so the GDN state should decay slowly; a small value means the head is mostly local, so a shorter memory is appropriate. The entropy statistic $e_h$ describes the shape of the same look-back distribution, telling whether probability mass is concentrated on a few past positions or spread across many.

We convert the average attention distance into a target exponential decay. If a head should preserve half of its information after roughly $d_h$ tokens, then the target decay magnitude is $|g_h|=\ln 2/d_h$ (Appendix~\ref{app:taylor}). At initialization we set $A_{\log}=0$, so the decay is controlled primarily by $dt_{\mathrm{bias}}$; with $a_{\mathrm{proj}}(x_t)\approx 0$, we have $g_h \approx -\operatorname{softplus}(dt_{\mathrm{bias},h})$, giving
\begin{equation}
dt_{\mathrm{bias},h}
:= \operatorname{softplus}^{-1}\!\left(\frac{\ln 2}{d_h}\right).
\end{equation}

We next convert entropy into a write-gate target. We normalize entropy within each layer into $c_h=1-(e_h-e_{\min})/(e_{\max}-e_{\min})$, so lower-entropy heads receive larger concentration scores, and map it to $\beta_h^\star=0.3+0.4c_h$ derived in Appendix~\ref{app:beta_derivation}. Let $\bar{u}_h=\mathbb{E}_{x_t\sim\mathcal{D}_{\mathrm{cal}}}[(b_{\mathrm{proj}}x_t)_h]$ be the average pre-sigmoid write logit for head $h$ on the calibration data. We initialize the corresponding row of $b_{\mathrm{proj}}$ by matching this average logit to $\operatorname{logit}(\beta_h^\star)$; Appendix~\ref{app:bproj_rescaling} gives the row-rescaling derivation:
\begin{equation}
(b_{\mathrm{proj}})_{h,:}\leftarrow
\frac{\operatorname{logit}(\beta_h^\star)}{\bar{u}_h}\,(b_{\mathrm{proj}})_{h,:}.
\end{equation}

Finally, we set the output scale. The value-side OLS factor $\sigma_h^\star$ is computed per head by matching the student value-path output to the teacher output in least squares \cite{bjorck1996least} (Appendix~\ref{app:ols_rescaling}). For the output gate, a random $g_{\mathrm{proj}}$ can dominate the layer output even when the recurrent state is reasonable, so we use an RMS-matched small gate. Let $\alpha_\ell^\star=\lambda\,\operatorname{RMS}(y_T^{(\ell)})/\operatorname{RMS}(\operatorname{SiLU}(W_V^{(\ell)}x))$ with $\lambda=0.01$ on calibration inputs (Appendix~\ref{app:gproj_init}). The corresponding initializations are
\begin{equation}
W_V^{(h)} \leftarrow \sigma_h^\star W_V^{(h)},\qquad
g_{\mathrm{proj}}^{(\ell)} \leftarrow \alpha_\ell^\star W_V^{(\ell)}.
\end{equation}
The first update transfers the Taylor-derived value amplitude match, while the second keeps the output gate in a low-contribution, high-gradient regime before layer-local alignment. 

\subsection{Phase 2: Per-Layer Gradient Alignment}

Analytical calibration sets the initial decay and gate scales, but a first-order approximation cannot determine the exact parameter directions needed to match the teacher. Our second stage therefore performs a brief \emph{layer-local} distillation step, in line with the broader use of hidden-state matching and knowledge distillation for converted sequence models \cite{hinton2015distilling,zhang2025lolcats,goldstein2025radlads}. For each converted layer $\ell$, we minimize
\begin{equation}
\mathcal{L}_{\mathrm{align}}^{(\ell)}
:=
\mathbb{E}_{x \sim \mathcal{D}_{\mathrm{cal}}}
\left\|
\hat{z}_{\mathrm{GDN}}^{(\ell)}(x; \theta_\ell)
-
z_T^{(\ell)}(x)
\right\|_2^2,
\end{equation}
where $\hat{z}_{\mathrm{GDN}}^{(\ell)}$ is the student GDN output at the same hidden-state input. We update both transferred projections and newly introduced recurrent parameters:
\begin{equation}
\theta_\ell =
\{W_Q,W_K,W_V,W_O,A_{\log},dt_{\mathrm{bias}},a_{\mathrm{proj}},b_{\mathrm{proj}},g_{\mathrm{proj}}\}.
\end{equation}

The two stages cover different errors. The Taylor-derived stage sets the cheap scale and gate choices analytically: it inherits value-side OLS calibration and replaces sequence-mixing rescaling with GDN-specific decay, write, and output-gate calibration. The alignment step handles the residual nonlinear mismatch that these low-order statistics cannot capture. It lets transferred projections adapt to recurrent rather than softmax dynamics and updates the \emph{direction} of the gates rather than merely their scale. As we show later, successful alignment does not dramatically increase the magnitude of $g_{\mathrm{proj}}$; instead it rotates these parameters into directions that produce teacher-like outputs.

\section{Experiments}
\label{sec:experiments}

\subsection{Experimental Setup}

We evaluate \methodname{} across four teacher settings:
\textbf{Qwen2.5-1.5B-Instruct} and
\textbf{Qwen2.5-3B-Instruct} \cite{qwen2024qwen25},
\textbf{Llama-3.2-3B-Instruct} \cite{meta2024llama32}, and
\textbf{Qwen3-8B} \cite{yang2025qwen3}.
For each teacher we replace a representative fraction of softmax attention layers with Gated DeltaNet (GDN) and compare five initialization strategies:
(i)~\textbf{Baseline}---copy teacher $Q/K/V/O$ projections, leave new recurrent parameters at their random default;
(ii)~\textbf{Zero-Gate}---same projection copy, set output gates $g_{\mathrm{proj}}=0$ so the recurrent branch contributes nothing at step~0;
(iii)~\textbf{Small-Gate}---set $g_{\mathrm{proj}}=\epsilon$ with $\epsilon=0.01$;
(iv)~\textbf{Taylor-Only}---Phase~1 analytical calibration (Section~\ref{sec:method}) without per-layer alignment; and
(v)~\textbf{\methodname}---full Phase~1 + Phase~2.
All variants use the same projection-transfer backbone: we first copy the teacher $Q/K/V/O$ weights following RADLADS \cite{goldstein2025radlads}, which is also the conversion pipeline used by GA-S2 after layer selection \cite{li2025klguided}. In those protocols, non-equivalent recurrent parameters are left to standard initialization or later alignment; our comparison keeps that backbone fixed and changes only how the new GDN dynamics are initialized. The baselines only change simple heuristics for the new recurrent parameters, Taylor-Only adds our Phase~1 analytical calibration, and \methodname{} further adds Phase~2 layer-local alignment. This setup isolates whether GDN-specific Taylor calibration places the recurrent block in a better initial dynamical regime.

We compare three retained-layer selection policies:
\textbf{Uniform} (evenly spaced softmax layers),
\textbf{AR (LM-PPL)} (the lightweight LM-PPL heuristic from \cite{li2025klguided}), and
\textbf{GA-S2} (the KL-guided selector proposed by \cite{li2025klguided}).

After zero-shot evaluation, all converted students use the same RADLADS-style two-stage distillation recipe: hidden-state alignment followed by KL-based distillation \cite{hinton2015distilling,goldstein2025radlads,li2025klguided}.
Short-context quality is measured with \textbf{perplexity~(PPL)} and twelve downstream tasks: \textbf{ARC-C/E} \cite{clark2018think}, \textbf{HellaSwag} \cite{zellers2019hellaswag}, \textbf{PIQA} \cite{bisk2020piqa}, \textbf{MMLU} \cite{hendrycks2021mmlu}, \textbf{OpenbookQA~(OBQA)} \cite{mihaylov2018openbookqa}, \textbf{ReArc~(RA)} \cite{chollet2019measure}, \textbf{WinoGrande~(WG)} \cite{sakaguchi2021winogrande}, \textbf{BoolQ} \cite{clark2019boolq}, \textbf{LAMBADA} \cite{paperno2016lambada}, \textbf{COPA} \cite{roemmele2011copa}, and \textbf{SciQ} \cite{welbl2017sciq}. The reported \textbf{Avg} is the macro average over these twelve downstream tasks; it excludes PPL and RULER.
We additionally report \textbf{RULER} \cite{hsieh2024ruler} as a lightweight long-context probe at initialization and across recovery checkpoints. In the current evaluation harness, \texttt{--tasks ruler} expands to 13 subtasks: eight \textbf{NIAH} variants (\texttt{niah\_single\_1/2/3}, \texttt{niah\_multikey\_1/2/3}, \texttt{niah\_multiquery}, and \texttt{niah\_multivalue}) plus five additional probes (\texttt{ruler\_vt}, \texttt{ruler\_cwe}, \texttt{ruler\_fwe}, \texttt{ruler\_qa\_squad}, and \texttt{ruler\_qa\_hotpot}).
Unless otherwise noted, the paper reports the aggregate \textbf{RULER} score over this bundled task set rather than breaking out the 13 subtasks individually.

\newcommand{\arpolicy}{AR (LM-PPL)}
\newcommand{\stageoneckpt}{100M {\scriptsize (Stage 1)}}
\newcommand{\stagetwockpt}{700M {\scriptsize (Stage 2)}}

\subsection{Zero-Shot Evaluation}
\label{sec:zero-shot}

Before spending any training tokens, we ask whether \methodname{} improves the initial Avg and RULER scores under the same projection-transfer setting used by recent linear-attention conversion methods \cite{wang2024mamba,goldstein2025radlads,li2025klguided}.
Table~\ref{tab:zero-shot} summarizes zero-shot Avg and RULER for the Uniform policy across 25\%, 50\%, and 75\% retained attention, together with the teacher reference.
Detailed task-level metrics and non-uniform policies are reported in Appendix~\ref{app:extended}, especially Appendix~\ref{app:main_detailed_tables}.
The teacher's own performance is included as a ceiling reference.

Table~\ref{tab:zero-shot} shows three clear trends. First, aggressive conversion remains difficult at initialization: with only 25\% or 50\% retained attention, all methods have very low RULER scores and the Baseline Avg is close to chance-level for several models. Second, simple output-gate stabilization is most useful when enough softmax layers remain. At 75\% retained attention, Zero-Gate and Small-Gate recover much of the short-context Avg and often improve RULER over the raw Baseline, showing that preventing an uncalibrated recurrent branch from perturbing the residual stream is already important. Third, the full \methodname{} initialization most consistently improves zero-shot Avg, especially at 50\% and 75\% retention, while RULER gains are more model- and budget-dependent. This suggests that layer-local alignment mainly improves the initial short-context Avg of the converted student, whereas long-context retrieval still requires retained attention or later recovery training.

\begin{table}[!htbp]
\centering
\scriptsize
\caption{Compact zero-shot summary for the Uniform retained-layer policy. Each entry reports \textbf{Avg/RULER}; full task-level results and non-uniform layer-selection policies are reported in Appendix~\ref{app:main_detailed_tables}. \textbf{Bold}: best non-teacher value within each model and budget row.}
\label{tab:zero-shot}
\renewcommand{\arraystretch}{0.9}
\setlength{\tabcolsep}{2.1pt}
\begin{tabular}{@{}llccccc>{\columncolor{blue!8}}c@{}}
\toprule
Model & Budget & Teacher & Baseline & Zero-Gate & Small-Gate & Taylor-Only & \methodname{} \\
\midrule
Qwen2.5-1.5B & 25\% & 64.8 / 86.2 & 31.2 / \textbf{0.2} & 31.3 / \textbf{0.2} & 31.4 / \textbf{0.2} & 31.9 / \textbf{0.2} & \textbf{34.8} / \textbf{0.2} \\
 & 50\% & 64.8 / 86.2 & 31.7 / \textbf{0.1} & 32.7 / \textbf{0.1} & 32.7 / \textbf{0.1} & 32.6 / \textbf{0.1} & \textbf{37.4} / \textbf{0.1} \\
 & 75\% & 64.8 / 86.2 & 48.6 / 2.7 & 53.9 / 10.1 & 53.7 / 10.3 & 53.9 / 10.1 & \textbf{56.2} / \textbf{14.5} \\
\addlinespace
Qwen2.5-3B & 25\% & 67.3 / 91.3 & 31.1 / 0.0 & \textbf{31.8} / 0.0 & \textbf{31.8} / 0.0 & 31.7 / 0.0 & 30.9 / \textbf{0.1} \\
 & 50\% & 67.3 / 91.3 & 31.3 / 0.0 & 31.3 / 0.0 & 31.4 / 0.0 & 31.3 / 0.0 & \textbf{36.3} / \textbf{0.4} \\
 & 75\% & 67.3 / 91.3 & 55.1 / 19.9 & 61.8 / 47.4 & \textbf{62.0} / 47.2 & 61.9 / 47.2 & 61.3 / \textbf{55.9} \\
\addlinespace
Llama-3.2-3B & 25\% & 65.6 / 89.6 & 31.2 / 0.0 & 33.1 / 0.0 & 33.2 / \textbf{0.1} & 33.1 / \textbf{0.1} & \textbf{34.2} / \textbf{0.1} \\
 & 50\% & 65.6 / 89.6 & 31.7 / 0.1 & 36.9 / 0.2 & 36.7 / 0.2 & 36.8 / 0.2 & \textbf{43.3} / \textbf{0.4} \\
 & 75\% & 65.6 / 89.6 & 33.9 / 0.2 & 58.3 / \textbf{31.0} & 58.6 / 30.9 & 58.5 / 30.9 & \textbf{60.0} / 24.0 \\
\addlinespace
Qwen3-8B & 25\% & 70.7 / 94.0 & 31.4 / \textbf{0.1} & 34.2 / \textbf{0.1} & 34.1 / 0.0 & 34.2 / \textbf{0.1} & \textbf{35.6} / \textbf{0.1} \\
 & 50\% & 70.7 / 94.0 & 36.2 / 0.2 & 45.1 / 0.3 & 44.9 / 0.3 & 45.0 / 0.3 & \textbf{49.0} / \textbf{0.7} \\
 & 75\% & 70.7 / 94.0 & 59.9 / 42.3 & 64.8 / 61.6 & 64.7 / 61.4 & 64.8 / 61.6 & \textbf{65.5} / \textbf{65.1} \\
\bottomrule
\end{tabular}
\end{table}

\subsection{Main Evaluation: Recovery at Token Checkpoints}
\label{sec:main-eval}

The zero-shot comparison establishes that \methodname{} provides a better starting point.
The next question is whether this advantage translates into faster recovery during downstream training: at which token checkpoint does the converted student reach teacher-like quality, and how much does better initialization accelerate restoration?

\begin{figure}[!htbp]
\centering
\begin{minipage}[t]{0.48\linewidth}
\centering
\textbf{(a) Llama-3.2-3B}\\
{\footnotesize Training dashboard}\\[2pt]
\includegraphics[width=\linewidth]{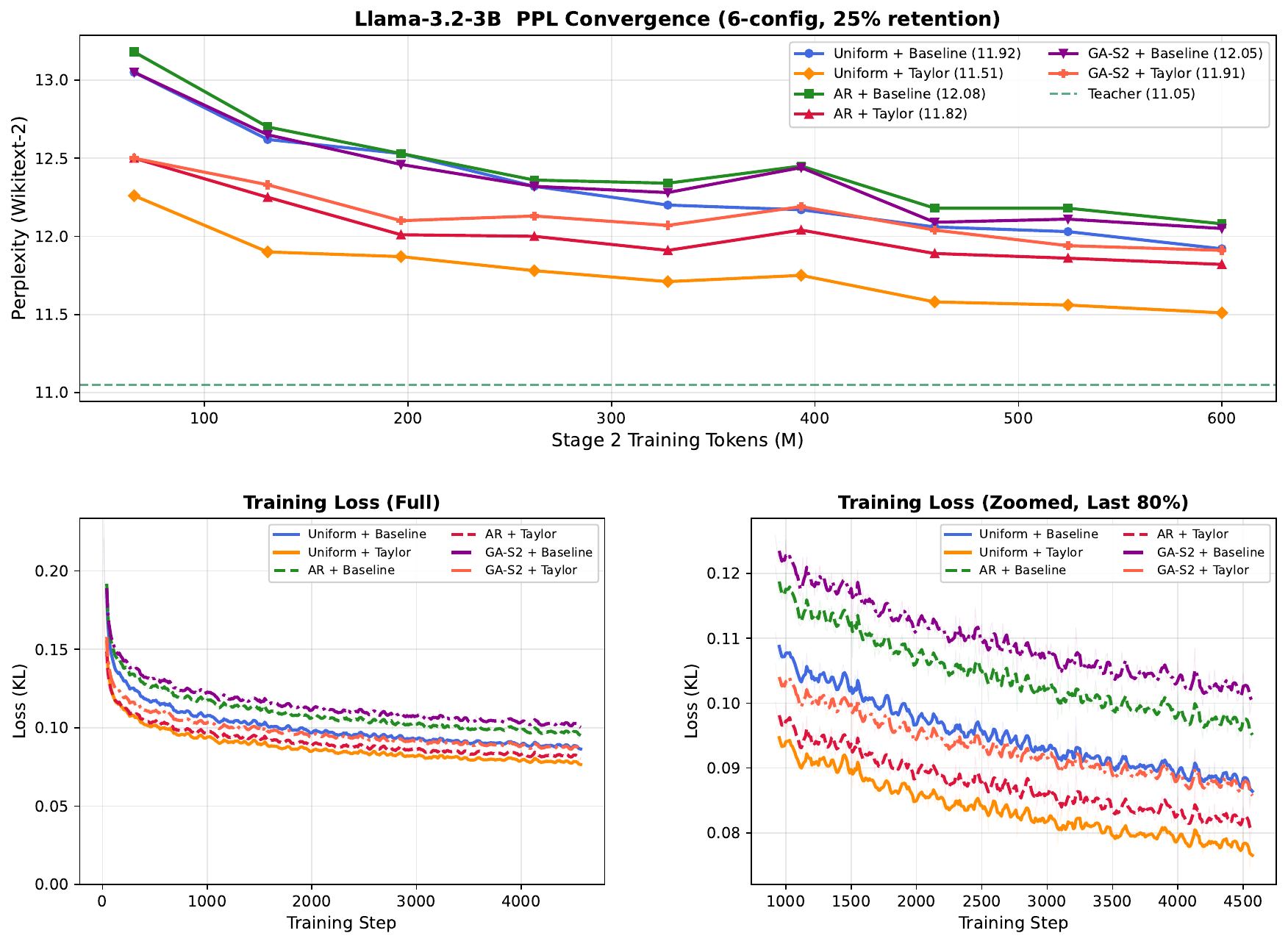}
\end{minipage}
\hfill
\begin{minipage}[t]{0.48\linewidth}
\centering
\textbf{(b) Qwen3-8B}\\
{\footnotesize Training dashboard}\\[2pt]
\includegraphics[width=\linewidth]{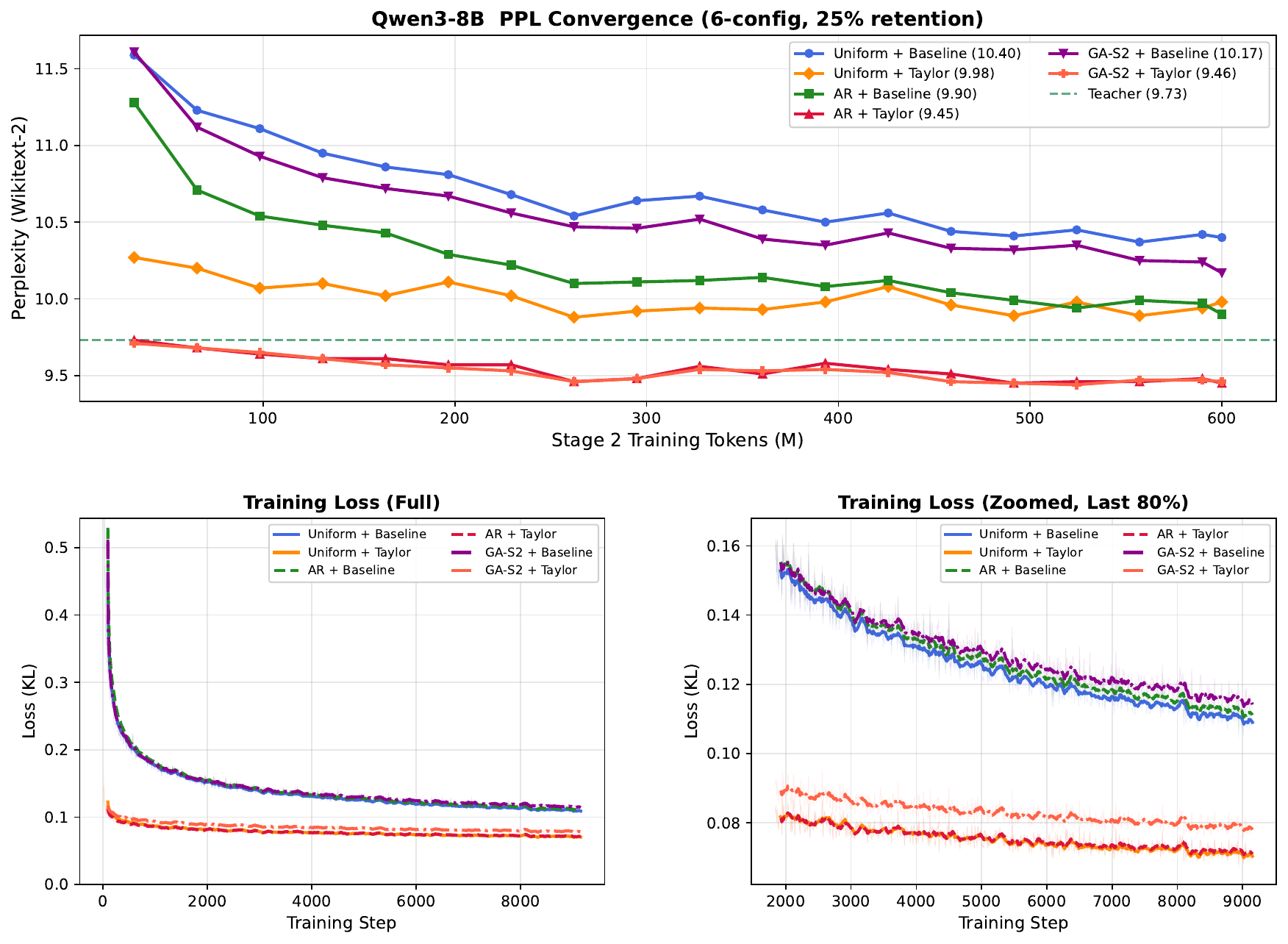}
\end{minipage}
\caption{Recovery dynamics under matched distillation training for the main teacher settings. Each subfigure shows the current training dashboard, summarizing the available recovery trajectories across configurations.}
\label{fig:recovery-curves}
\vspace{-0.4cm}
\end{figure}

Table~\ref{tab:recovery-summary} summarizes checkpoint recovery for all teacher settings, comparing Baseline and \methodname{} across retained-layer policies with Avg and RULER.
Figure~\ref{fig:recovery-curves} then shows training dashboards for the main recovery runs, tracking PPL and optimization losses over the consumed token budget.
Detailed task-level metrics are reported in Appendix~\ref{app:extended}, especially Appendix~\ref{app:main_detailed_tables}.

\begin{table}[!htbp]
\centering
\scriptsize
\caption{Compact recovery summary for all teacher settings. Each entry reports \textbf{Avg/RULER}; detailed task-level metrics are reported in Appendix~\ref{app:main_detailed_tables}. \textbf{Bold}: better value between Baseline and \methodname{} within each checkpoint.}
\label{tab:recovery-summary}
\renewcommand{\arraystretch}{0.9}
\setlength{\tabcolsep}{2.0pt}
\begin{tabular}{@{}llcc>{\columncolor{blue!8}}cc>{\columncolor{blue!8}}c@{}}
\toprule
Model & Selection & Teacher & \multicolumn{2}{c}{100M {\tiny (Stage 1)}} & \multicolumn{2}{c}{700M {\tiny (Stage 2)}} \\
\cmidrule(lr){4-5}\cmidrule(l){6-7}
 &  &  & Baseline & \methodname{} & Baseline & \methodname{} \\
\midrule
Qwen2.5-1.5B & Uniform & 64.8 / 86.2 & 37.2 / 0.6 & \textbf{62.6} / \textbf{7.7} & 61.4 / 39.9 & \textbf{64.4} / \textbf{59.8} \\
 & \arpolicy & 64.8 / 86.2 & 44.7 / 1.0 & \textbf{61.1} / \textbf{5.0} & 61.4 / 16.8 & \textbf{63.6} / \textbf{38.0} \\
 & GA-S2 & 64.8 / 86.2 & 43.6 / 0.5 & \textbf{62.0} / \textbf{7.1} & 60.3 / 27.0 & \textbf{63.8} / \textbf{47.6} \\
\addlinespace
Qwen2.5-3B & Uniform & 67.3 / 91.3 & 60.5 / 26.3 & \textbf{60.8} / \textbf{29.2} & 65.8 / 63.8 & \textbf{66.0} / \textbf{65.7} \\
 & \arpolicy & 67.3 / 91.3 & \textbf{62.1} / 11.5 & \textbf{62.1} / \textbf{11.7} & 65.8 / 56.1 & \textbf{66.1} / \textbf{57.9} \\
 & GA-S2 & 67.3 / 91.3 & 52.6 / \textbf{6.6} & \textbf{60.4} / 4.9 & 65.7 / 59.8 & \textbf{65.9} / \textbf{70.5} \\
\addlinespace
Llama-3.2-3B & Uniform & 65.6 / 89.6 & 59.7 / 5.7 & \textbf{61.3} / \textbf{6.6} & 64.3 / 60.0 & \textbf{64.8} / \textbf{63.4} \\
 & \arpolicy & 65.6 / 89.6 & 58.5 / 4.6 & \textbf{60.1} / \textbf{5.4} & 63.8 / 46.4 & \textbf{64.0} / \textbf{47.3} \\
 & GA-S2 & 65.6 / 89.6 & 57.8 / \textbf{5.0} & \textbf{59.5} / 4.2 & 62.3 / 41.1 & \textbf{63.9} / \textbf{46.5} \\
\addlinespace
Qwen3-8B & Uniform & 70.7 / 94.0 & 49.4 / 0.1 & \textbf{66.5} / \textbf{9.2} & 69.4 / 60.4 & \textbf{70.0} / \textbf{76.7} \\
 & \arpolicy & 70.7 / 94.0 & 44.7 / 0.2 & \textbf{67.6} / \textbf{15.4} & 69.2 / 62.2 & \textbf{70.1} / \textbf{69.8} \\
 & GA-S2 & 70.7 / 94.0 & 39.8 / 0.1 & \textbf{67.4} / \textbf{11.0} & 68.9 / \textbf{66.1} & \textbf{70.1} / 57.2 \\
\bottomrule
\end{tabular}
\end{table}

Table~\ref{tab:recovery-summary} shows that the largest benefit of \methodname{} appears early in recovery, before Stage~2 has had enough tokens to repair a poor initialization. At 100M tokens, the gains are especially large for the harder settings: Qwen2.5-1.5B improves from 37.2 to 62.6 Avg under Uniform selection, and Qwen3-8B improves from 39.8--49.4 to 66.5--67.4 Avg across the three retained-layer policies. Llama-3.2-3B has a smaller Avg gap because the baseline already recovers reasonably after Stage~1, but \methodname{} still gives consistent short-context gains.

By 700M tokens, the Avg gap narrows because distillation has largely restored short-context behavior for both initializations. \methodname{} still gives the best Avg in every 700M row and often improves RULER as well, although long-context retrieval remains below the teacher. This suggests that better initialization mainly reduces the amount of training needed to approach teacher-level short-context quality, while full long-context restoration is a slower process.

The training dashboards in Figure~\ref{fig:recovery-curves} are consistent with this interpretation. PPL and loss decrease smoothly for both teachers, and we do not observe instability that would explain the gap by a pathological baseline run. Instead, the \methodname{} curves start from a lower-loss state and reach the near-teacher PPL region with fewer tokens, while the final curves become more tightly clustered after extended distillation. Together with Table~\ref{tab:recovery-summary}, this indicates that initialization quality mainly changes recovery speed under the fixed downstream training objective.

\subsection{Long-Context Restoration}
\label{sec:long-context}

RULER \cite{hsieh2024ruler} recovers more slowly than PPL and short-context Avg. At 700M tokens, Llama-3.2-3B-Instruct nearly matches the teacher on Avg (64.8 vs.\ 65.6), but still trails on RULER (63.4 vs.\ 89.6). Qwen3-8B shows the same pattern: the best reported Stage-2 RULER score is 76.7, still below the teacher at 94.0. Thus 700M tokens can recover much of the short-context behavior, but it does not fully restore long-context retrieval.

\begin{wrapfigure}{r}{0.48\linewidth}
    \vspace{-10pt}
    \centering
    \includegraphics[width=\linewidth]{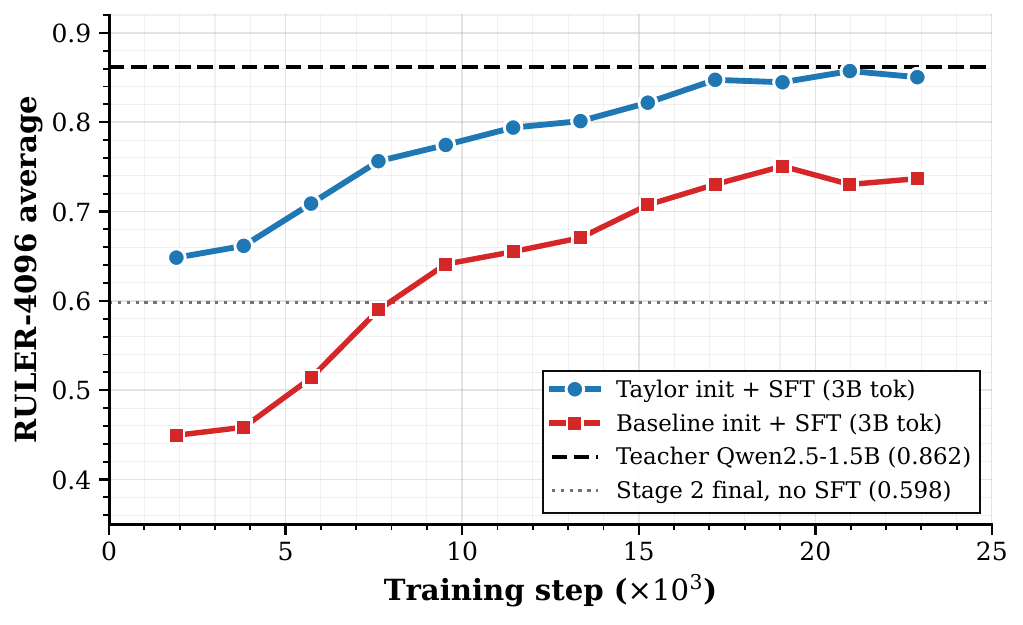}
    \caption{RULER trajectory for the 1.5B setting during extended SFTMix v0.2 training.}
    \label{fig:ruler-trajectory}
    \vspace{-10pt}
\end{wrapfigure}

Figure~\ref{fig:ruler-trajectory} shows the longer-horizon behavior for the 1.5B setting. We continue training on the SFTMix v0.2 corpus for more than 20k optimization steps, rather than stopping at the short recovery schedule used in Table~\ref{tab:recovery-summary}. Under this data and longer training budget, RULER continues to improve and reaches the teacher range in this run, indicating that the Stage-2 long-context gap can be reduced with additional recovery training.

The conclusion is that \methodname{} improves the starting point and early recovery, but RULER is governed by a slower timescale than PPL or short-context Avg. In our 1.5B run, restoring retrieval behavior requires SFTMix v0.2 and more than 20k optimization steps, which is longer than the short recovery schedule used in Table~\ref{tab:recovery-summary}.

\subsection{Ablations}
\label{sec:ablation}

\subsubsection{The Importance of Analytical Calibration and Local Alignment}

We ablate the two stages of \methodname{} on Qwen2.5-3B-Instruct \cite{qwen2024qwen25} with the Uniform policy.
Table~\ref{tab:ablation-components} compares five variants under the same hybrid budget.

\begin{table}[!htbp]
\centering
\small
\caption{Component ablation on Qwen2.5-3B-Instruct (Uniform). Avg denotes the short-context NLU average from this ablation sweep; MMLU is unavailable in this run.}
\label{tab:ablation-components}
\setlength{\tabcolsep}{5pt}
\begin{tabular}{@{}lccl@{}}
\toprule
Variant & PPL$\downarrow$ & Avg & Description \\
\midrule
Baseline & 37337.3 & 30.9 & projection copy only \\
Zero-Gate & 22469.0 & 30.7 & output-path stabilization \\
Taylor-Only & 22470.6 & 30.9 & Phase~1 only \\
Alignment-Only & 2015.9 & 31.4 & Phase~2 only, no analytical calibration \\
\methodname & \textbf{424.1} & \textbf{32.1} & Phase~1 + Phase~2 \\
\bottomrule
\end{tabular}
\end{table}

Phase~2 alignment is essential. Baseline starts at PPL 37337.3, while Zero-Gate and Taylor-Only remain around 22470, so Phase~1 alone is not enough. Alignment-Only lowers PPL to 2015.9, and the full \methodname{} reaches 424.1 with the best Avg, 32.1. This shows that Taylor calibration is most useful as a better starting point for local alignment, not as a standalone replacement for it.

\begin{wrapfigure}{l}{0.52\linewidth}
    \centering
    \includegraphics[width=\linewidth]{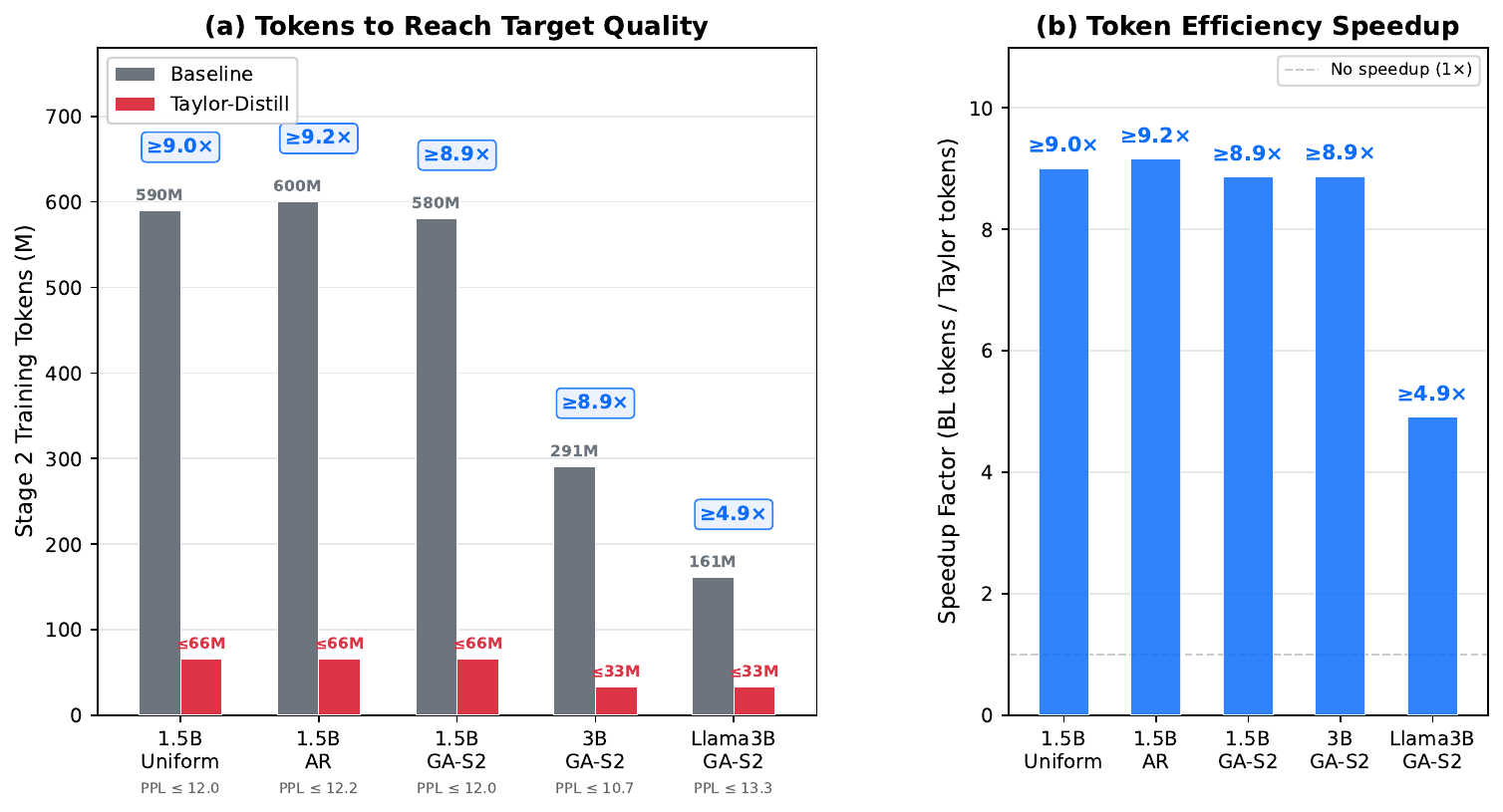}
    \caption{Token-budget comparison at matched target quality. For each representative run, panel (a) reports the Stage-2 training tokens required for Baseline and \methodname{} to reach the same target performance, and panel (b) reports the resulting speedup factor. Across these runs, \methodname{} reduces the required budget by 4.9$\times$--9.2$\times$.}
    \label{fig:token-budget}
    \vspace{-0.5cm}
\end{wrapfigure}

Appendix~\ref{app:gate_stats} further analyzes the initialized gate and half-life statistics.

\subsubsection{How Many Training Tokens Does Recovery Really Need?}

A practical question is how much Stage-2 distillation is needed after initialization.
Figure~\ref{fig:token-budget} compares the number of training tokens required for Baseline and \methodname{} to reach the same target quality in several representative settings.
Across the representative runs in Figure~\ref{fig:token-budget}, \methodname{} reaches the target with fewer tokens, yielding speedups from 4.9$\times$ to 9.2$\times$. This suggests that a large fraction of the recovery budget is spent moving the converted model out of a poor initial regime.

Appendix~\ref{app:discussion} discusses limitations and future directions, including larger-model support, architecture-specific calibration, and finer-grained head-wise conversion.

\section{Conclusion}
\label{sec:conclusion}

We presented \methodname, a lightweight initialization method for converting pretrained Transformers into GDN hybrid students. Simply copying the teacher projections leaves the new recurrent block with poorly calibrated memory timescales and gates, so downstream distillation has to spend many tokens repairing the mismatch. \methodname{} addresses this by combining Taylor-derived calibration with a short layer-local alignment stage before global distillation. Across the evaluated teachers and retained-layer policies, this changes failed zero-shot hybrids into usable students, improves the worst initial PPL by 88$\times$, and reduces the Stage-2 token budget needed to reach matched quality by 4.9$\times$--9.2$\times$. These results point to initialization as an important part of architecture transfer, alongside layer selection and distillation. Extending the same calibration view to other linear-attention families is a natural direction for future work.

\clearpage
\bibliography{references}
\bibliographystyle{plainnat}

\clearpage
\appendix
\section{Additional Preliminaries and Derivations}
\label{app:calibration}

\subsection{Hybrid Attention Architectures}
\label{app:hybrid_architectures}

We study \emph{hybrid} models that keep a subset of layers as full softmax attention and replace the remainder with GDN \cite{lieber2024jamba,jamba2024jamba15,yang2024gdn}. Let
\begin{equation}
\mathcal{S}_{\mathrm{soft}} \cup \mathcal{S}_{\mathrm{gdn}} = \{1,\ldots,L\},\qquad
\mathcal{S}_{\mathrm{soft}} \cap \mathcal{S}_{\mathrm{gdn}} = \varnothing.
\end{equation}
Layer selection is an important problem in its own right \cite{li2025klguided,hoshino2025rad}, but it is orthogonal to our question. We assume that a set of layers has already been chosen for replacement, and ask: \emph{how should the newly introduced GDN layers be initialized so that distillation starts from calibrated decay, write, and output-gate scales?}

\subsection{Decay Calibration from Half-Life Matching}
\label{app:taylor}

Our decay calibration uses a simple half-life argument to map teacher softmax-attention statistics \cite{vaswani2017attention} into the recurrent decay parameterization of GDN \cite{yang2024gdn}. Let $\rho_h = \exp(g_h)$ denote the per-step multiplicative decay for head $h$. If the teacher's average attention distance is $d_h$, we want the recurrent memory to decay by half after roughly $d_h$ tokens:
\begin{equation}
\rho_h^{d_h} = \frac{1}{2}.
\end{equation}
Solving for $g_h$ gives
\begin{equation}
g_h = -\frac{\ln 2}{d_h}.
\end{equation}
At initialization we set $A_{\log}=0$ and assume the input-dependent term $a_{\mathrm{proj}}(x_t)$ is small, so the GDN decay becomes
\begin{equation}
g_h \approx -\operatorname{softplus}(dt_{\mathrm{bias},h}).
\end{equation}
This immediately yields the closed-form initialization
\begin{equation}
dt_{\mathrm{bias},h} = \operatorname{softplus}^{-1}\!\left(\frac{\ln 2}{d_h}\right).
\end{equation}

\subsection{Deriving the Entropy-to-Write-Gate Mapping}
\label{app:beta_derivation}

We now justify the affine write-gate target used in the main text. The goal is to turn the teacher attention distribution into a stable GDN write-gate initialization \cite{yang2024gdn}:
\begin{equation}
\beta_h^\star = 0.3 + 0.4 c_h,
\end{equation}
where $c_h \in [0,1]$ is the normalized concentration score derived from attention entropy.
The point of this derivation is not to recover an exact nonlinear teacher-to-student transfer law, but to obtain a stable first-order rule that is appropriate for initialization near the neutral regime where the converted layer starts.

Because the decay timescale is already handled by $g_h$, the role of the write gate is only to modulate update sharpness around a neutral operating point. It is therefore natural to require that a head with median concentration initializes near a balanced write gate:
\begin{equation}
c_h = \frac{1}{2}
\quad \Longrightarrow \quad
\beta_h^\star = \frac{1}{2}.
\end{equation}
To avoid saturating the sigmoid parameterization in the GDN update, we work on the logit scale and introduce a bounded target
\begin{equation}
z_h^\star = \gamma \left(c_h - \frac{1}{2}\right),
\qquad
\beta_h^\star = \sigma(z_h^\star).
\end{equation}
We choose $\gamma$ so that even the most diffuse and most concentrated heads remain in the high-gradient regime of the sigmoid. Imposing
\begin{equation}
|z_h^\star| \le 0.8
\end{equation}
keeps
\begin{equation}
\sigma'(z_h^\star)
=
\sigma(z_h^\star)\bigl(1-\sigma(z_h^\star)\bigr)
\ge
\sigma(0.8)\bigl(1-\sigma(0.8)\bigr)
\approx 0.214,
\end{equation}
which is still about $86\%$ of the maximum slope $\sigma'(0)=0.25$. Since $c_h \in [0,1]$ implies $\left|c_h - \frac{1}{2}\right| \le \frac{1}{2}$, the largest such linear logit map uses
\begin{equation}
\gamma = 1.6.
\end{equation}
This gives the bounded nonlinear target
\begin{equation}
\beta_h^\star
=
\sigma\!\left(1.6\left(c_h - \frac{1}{2}\right)\right).
\end{equation}
Finally, we take the first-order Taylor expansion around the neutral point $c_h=\frac{1}{2}$:
\begin{equation}
\beta_h^\star
\approx
\sigma(0) + \sigma'(0)\,1.6\left(c_h - \frac{1}{2}\right)
=
\frac{1}{2} + 0.4\left(c_h - \frac{1}{2}\right)
=
0.3 + 0.4 c_h.
\end{equation}
Thus the affine rule used in the main text can be viewed as the first-order approximation of a bounded logit parameterization that keeps initialization away from both under-writing and sigmoid saturation.

\subsection{Matching the Write-Gate Logit by Row Rescaling}
\label{app:bproj_rescaling}

The previous subsection gives a target GDN write gate $\beta_h^\star$ \cite{yang2024gdn} and therefore a target logit
\begin{equation}
z_h^\star = \operatorname{logit}(\beta_h^\star).
\end{equation}
In Phase~1 we do not replace row $h$ of $b_{\mathrm{proj}}$ with a freshly optimized vector. Instead, we preserve the module-default direction and solve only for a scalar rescaling. Writing the default row as $\tilde{b}_h \in \mathbb{R}^{d_{\mathrm{in}}}$, we restrict the initialized row to
\begin{equation}
b_{\mathrm{proj}}[h,:] = s_h \tilde{b}_h.
\end{equation}

To convert the target logit into a row scale, we use a simple proxy for the row's typical preactivation magnitude under unit-RMS inputs:
\begin{equation}
m_h := \sqrt{d_{\mathrm{in}}}\,\operatorname{mean}\!\left(|\tilde{b}_h|\right).
\end{equation}
This proxy is exact when all coordinates of $\tilde{b}_h$ have the same magnitude and serves as a cheap row-wise approximation to the row's Euclidean scale more generally. We then choose $s_h$ so that the signed typical preactivation matches the target logit:
\begin{equation}
s_h^\star
:=
\arg\min_s \left(s\,m_h - z_h^\star\right)^2.
\end{equation}
Because $m_h > 0$, the minimizer is immediate:
\begin{equation}
s_h^\star
=
\frac{z_h^\star}
{\sqrt{d_{\mathrm{in}}}\,\operatorname{mean}\!\left(|\tilde{b}_h|\right)}.
\end{equation}
Substituting $z_h^\star = \operatorname{logit}(\beta_h^\star)$ and replacing $\tilde{b}_h$ by the current module-default row yields exactly the update used in the main text:
\begin{equation}
b_{\mathrm{proj}}[h,:]
\leftarrow
\frac{\operatorname{logit}(\beta_h^\star)}
{\sqrt{d_{\mathrm{in}}}\,\operatorname{mean}\!\left(|b_{\mathrm{proj}}[h,:]|\right)}
\; b_{\mathrm{proj}}[h,:].
\end{equation}
Thus the row-rescaling rule is the one-scalar least-squares match that preserves the default row direction while placing the write-gate preactivation in the teacher-informed logit regime.

\subsection{Closed-Form Value-Side OLS Rescaling}
\label{app:ols_rescaling}

For a fixed head $h$, flatten the indices $(b,t,d)$ into a single index $i$ and write
\begin{equation}
u_i := y_{T,b,t,h,d},
\qquad
v_i := y_{S,b,t,h,d}.
\end{equation}
The value-side calibration solves
\begin{equation}
\sigma_h^\star
=
\arg\min_{\sigma}
\sum_i \left(u_i - \sigma v_i\right)^2.
\end{equation}
Expanding the quadratic gives
\begin{equation}
J(\sigma)
=
\sum_i u_i^2
- 2\sigma \sum_i u_i v_i
+ \sigma^2 \sum_i v_i^2.
\end{equation}
Differentiating and setting the derivative to zero yields
\begin{equation}
\frac{dJ}{d\sigma}
=
-2 \sum_i u_i v_i
+ 2\sigma \sum_i v_i^2
=
0,
\end{equation}
so, provided the student context is not identically zero,
\begin{equation}
\sigma_h^\star
=
\frac{\sum_i u_i v_i}{\sum_i v_i^2}
=
\frac{\sum_{b,t,d} y_{T,b,t,h,d}\,y_{S,b,t,h,d}}
{\sum_{b,t,d} y_{S,b,t,h,d}^2}.
\end{equation}
This is the standard one-dimensional ordinary least-squares solution \cite{bjorck1996least}: it rescales the student context along its current direction so that it best matches the teacher in squared error. The clipping step mentioned in the main text is an implementation safeguard applied after this closed-form solution is computed.

\subsection{RMS-Matched Output-Gate Initialization}
\label{app:gproj_init}

After the OLS step, the value projection already defines a teacher-aligned feature direction. Because GDN further modulates the recurrent output with an output gate \cite{yang2024gdn}, we restrict the initial output-gate projection to the one-parameter family
\begin{equation}
W_g = \alpha_\ell W_V.
\end{equation}
Let
\begin{equation}
\phi_t^{(\ell)} := \operatorname{SiLU}(W_V x_t)
\end{equation}
be the unscaled gate-feature prior on the calibration set, and let
\begin{equation}
r_T^{(\ell)} := \operatorname{RMS}\!\left(y_T^{(\ell)}\right)
\end{equation}
denote the teacher pre-$W_O$ output scale. We want the initial gate contribution to be only a small fraction $\lambda$ of that scale. A simple scalar matching objective is therefore
\begin{equation}
\alpha_\ell^\star
:=
\arg\min_{\alpha \ge 0}
\left(
\operatorname{RMS}\!\left(\alpha \phi^{(\ell)}\right)
- \lambda r_T^{(\ell)}
\right)^2.
\end{equation}
Using the positive homogeneity of RMS for $\alpha \ge 0$,
\begin{equation}
\operatorname{RMS}\!\left(\alpha \phi^{(\ell)}\right)
=
\alpha\,\operatorname{RMS}\!\left(\phi^{(\ell)}\right),
\end{equation}
the minimizer is
\begin{equation}
\alpha_\ell^\star
=
\lambda\,
\frac{\operatorname{RMS}\!\left(y_T^{(\ell)}\right)}
{\operatorname{RMS}\!\left(\operatorname{SiLU}(W_V x)\right)}.
\end{equation}
This gives the formula used in the main text. In the actual parameterization the gate is evaluated as $\operatorname{SiLU}(W_g x) = \operatorname{SiLU}(\alpha_\ell W_V x)$, so the derivation above should be read as a calibration proxy rather than a strict identity. The small damping factor $\lambda = 0.01$ keeps the initialized gate in a near-linear, low-contribution regime, which is precisely the regime where this proxy is most appropriate.

\section{Related Work}
\label{sec:related}

\paragraph{Efficient hybrid sequence models.}
Our work is motivated by the growing use of efficient sequence mixers in large language models. Production-scale hybrid and linear architectures now include Jamba, Jamba-1.5, MiniMax-01, Falcon-H1, and Kimi Linear \cite{lieber2024jamba,jamba2024jamba15,minimax2025minimax01,falcon2025h1,kimi2025linear}. This trend continues in the Qwen3.5 release, whose model card highlights a Gated DeltaNet-based efficient hybrid architecture \cite{qwen2026qwen35}. A related efficiency direction compresses softmax attention itself, for example by using covariance-aware and rank-enhanced decomposition to enable multi-head latent attention \cite{zhou2026care}. These systems differ in their exact mixers, but they make the same architectural bet: many softmax attention layers can be replaced or compressed to reduce KV-cache cost and improve decoding throughput while preserving competitive quality.

\paragraph{Converting pretrained Transformers.}
A second line of work asks how to obtain such efficient models without pretraining them from scratch. Early work studied swap-then-finetune conversion \cite{kasai2021finetuning}. More recent methods, including SUPRA, Mamba in the Llama, LoLCATs, Liger, Llamba, and RADLADS, improve transfer through attention transfer, hidden-state alignment, low-rank finetuning, or progressive distillation \cite{mercat2024linearizing,wang2024mamba,zhang2025lolcats,lan2025liger,bick2025llamba,goldstein2025radlads}. Among these, RADLADS is the closest training protocol to our setup: it first transfers attention-related weights such as $W_Q,W_K,W_V,W_O$, initializes non-equivalent recurrent parameters with the standard student initialization, and then applies hidden-state alignment followed by KL distillation \cite{goldstein2025radlads}. GA-S2 builds on this RADLADS-style pipeline for GDN hybrids, using teacher-initialized $Q/K/V/O$ projections while focusing its contribution on selecting which layers should remain softmax attention; its other GDN-specific parameters are not calibrated from teacher attention statistics \cite{li2025klguided}. These works show that pretrained Transformer knowledge can be preserved during architectural conversion, but they mostly treat the initial setting of newly introduced recurrent dynamics as part of the default conversion recipe.

\paragraph{Layer selection in hybrid models.}
For hybrid conversion, the choice of \emph{which} layers remain softmax attention is also important. Redundancy-aware replacement and KL-guided layer selection show that layer choice can strongly affect long-context behavior and recovery quality \cite{hoshino2025rad,li2025klguided}. This question is complementary to ours. We assume that a replacement set has already been chosen and focus on the next step: once a layer is converted, how should the new recurrent dynamics be initialized?

\paragraph{Initialization and local transfer.}
\methodname{} is most directly related to transfer schemes that reuse teacher projections or perform layer-local matching. Mamba in the Llama explicitly initializes recurrent modules from attention projections \cite{wang2024mamba}, and LoLCATs uses local attention matching before global finetuning \cite{zhang2025lolcats}. We build on this view but shift the emphasis to the parameters that are \emph{not} inherited from the teacher. In GDN conversion, decay, write, and output-gate parameters determine the initial dynamical regime and can dominate zero-shot quality. Our Taylor view of softmax attention is therefore used as a calibration tool for these new GDN parameters, rather than as a literal proposal to reuse the same linear surrogate.

\section{Limitations and Future Work}
\label{app:discussion}

\methodname{} is a lightweight initialization method, not a complete solution to Transformer-to-recurrent conversion. Its role is to give the converted GDN layers a better starting point before full distillation, so several design choices remain deliberately simple. We summarize the main limitations below and describe concrete directions that could turn the initialization procedure into a more complete conversion recipe.

\paragraph{Larger Model Support}
The experiments in the main paper focus on teacher settings for which we can run the full conversion, recovery, and evaluation pipeline end to end. Extending the same protocol to larger teachers such as Qwen3-30B-A3B \cite{yang2025qwen3}, GLM \cite{glm2024chatglm}, and MiniMax-style full-attention models \cite{minimax2025minimax01} requires additional system support: memory-efficient teacher forward passes, distributed evaluation, and matched recovery schedules across retained-attention policies. We therefore keep the quantitative claims in this paper to the completed teacher settings, and treat larger-model support as an engineering extension of the current framework rather than an evaluated result. The same initialization, recovery, and long-context evaluation protocol can be applied once the distributed conversion stack supports these teachers.

\paragraph{Architecture specificity.}
We focus on Gated DeltaNet \cite{yang2024gdn} because it exposes clear recurrent quantities: decay, write gate, value pathway, and output gate. This makes the analysis clean, but the same formulas should not be treated as architecture-independent recipes. GLA \cite{yang2024gla}, Mamba-like state-space layers \cite{gu2024mamba,dao2024ssm}, KDA \cite{kimi2025linear}, and other efficient sequence models use different parameterizations and different mechanisms for memory update and input-dependent gating. The value-side least-squares step should transfer broadly because it only matches output amplitude, but the decay and gate calibration should be rederived for each architecture. A useful next step is to identify the small set of dynamical quantities each architecture needs at initialization, then map teacher attention statistics to those quantities in the architecture's own parameterization. This would test whether the broader idea behind \methodname{} is general, rather than only the particular GDN formulas.

\paragraph{Layer-wise rather than head-wise hybridization.}
We choose softmax layers at the layer level, following the layer-selection framing used by recent hybrid conversion work \cite{hoshino2025rad,li2025klguided}. This keeps the experimental setup controlled and makes the retained-attention budget easy to compare, but it ignores an important source of structure inside a Transformer layer: different heads can serve very different roles. Some heads may perform retrieval or copying over long ranges, while others mostly aggregate local context or syntactic features. Converting all heads in a selected layer to GDN therefore may be too coarse, especially for models whose long-context behavior depends on a small number of specialized attention heads. A natural next step is head-wise conversion: keep retrieval-heavy heads as softmax, convert more local or diffuse heads to GDN, and initialize each converted head using the same distance, entropy, and value-scale statistics used by \methodname{}. This would also make the retained-attention budget more fine-grained, since a model could preserve only the heads that actually need quadratic attention.

\paragraph{Heuristic calibration rules.}
Several calibration choices are still heuristic, including the half-life rule, entropy-to-write-gate map, clipping ranges, and output-gate scale. These choices are designed to be simple, cheap, and stable near initialization, and Appendix~\ref{app:gate_stats} shows that they produce the intended memory-timescale behavior. However, they are not globally optimal for the nonlinear GDN recurrence, and they do not account for interactions among decay, write strength, key normalization, and the output gate. For example, the same average attention distance can arise from different attention shapes, and the same entropy can correspond to different retrieval patterns. Future work could replace these hand-designed maps with a constrained matching objective that directly solves for decay and gate parameters from teacher statistics. A direct extension would fit the recurrent parameters by matching short teacher attention rollouts, while constraining the learned gates to remain in a stable range before downstream distillation.

\paragraph{Calibration and evaluation scope.}
Our calibration corpus is small, and our layer-local alignment stage is short. This is intentional: the method is meant to be an initialization step that costs little compared with full recovery training. Still, this leaves several open questions about sensitivity. Preliminary sweeps show useful trends for calibration size, alignment steps, and learning rate, but we have not fully tested domain mismatch, longer calibration sequences, much larger calibration sets, or more aggressive local optimization schedules. The current experiments also emphasize recovery checkpoints rather than exhaustive multi-seed sweeps, so they are best read as evidence about the effectiveness and token efficiency of the initialization rather than a complete variance study. Future work should run broader sensitivity analyses across domains, sequence lengths, seeds, and teacher families, and should measure whether the same initialization advantages persist after longer recovery training.

\paragraph{Downstream distillation dependence.}
\methodname{} improves the starting point, but it does not remove the need for downstream distillation \cite{hinton2015distilling,goldstein2025radlads,li2025klguided}. The converted student still has a different sequence mixer from the teacher, so global training is needed to repair cross-layer interactions and adapt the residual stream to the new recurrent blocks. The final quality therefore depends on layer selection, retained-softmax budget, distillation loss, optimizer settings, and the number of recovery tokens. In this paper we keep those components mostly fixed so that the effect of initialization is easier to isolate. A more complete conversion recipe could jointly choose which layers or heads to retain, initialize the converted recurrent modules, and schedule the distillation losses based on how quickly each part of the model recovers. Another useful direction is adaptive token allocation: spend more recovery tokens on conversions whose local alignment loss or early validation PPL indicates a harder mismatch.

\paragraph{Broader impact.}
This work reduces the compute and token budget needed for hybrid conversion. That can make efficient long-context models easier to build and evaluate, especially for groups that cannot afford full pretraining of a new architecture. The same efficiency gain also inherits the usual dual-use risks of language-model research: cheaper adaptation can support beneficial applications, but it can also lower the cost of producing capable models for harmful uses. We do not release new models or datasets here, and the method should be viewed as an architecture-transfer tool rather than a deployment-ready safety mechanism. Any deployed student model would still require the same safety evaluation, data governance, and misuse analysis as other language models. Future work on efficient conversion should therefore report not only quality and compute savings, but also the evaluation procedures used before release.

\section{Extended Experimental Results}
\label{app:extended}

This appendix collects supplementary empirical material referenced in Sections~\ref{sec:experiments} and \ref{sec:ablation}. It reports additional teacher settings and the gate-statistics diagnostics used to interpret the initialization behavior.

\subsection{Detailed Experimental Results}
\label{app:main_detailed_tables}
\label{app:qwen25_results}

This section provides the detailed task-level results that supplement the compact main-paper summaries in Tables~\ref{tab:zero-shot} and~\ref{tab:recovery-summary}. The teacher settings use Qwen2.5 \cite{qwen2024qwen25}, Llama-3.2 \cite{meta2024llama32}, and Qwen3 \cite{yang2025qwen3}. The short-context suite includes ARC-C/E \cite{clark2018think}, HellaSwag \cite{zellers2019hellaswag}, PIQA \cite{bisk2020piqa}, MMLU \cite{hendrycks2021mmlu}, OpenbookQA \cite{mihaylov2018openbookqa}, ReArc \cite{chollet2019measure}, WinoGrande \cite{sakaguchi2021winogrande}, BoolQ \cite{clark2019boolq}, LAMBADA \cite{paperno2016lambada}, COPA \cite{roemmele2011copa}, and SciQ \cite{welbl2017sciq}, while RULER is reported as the long-context probe \cite{hsieh2024ruler}. The organization matches the main experimental flow: zero-shot evaluation for all four teacher settings, recovery dashboards for the two Qwen2.5 settings, and checkpoint recovery for all four teacher settings.

\paragraph{Zero-shot details.}
Table~\ref{tab:appendix-detailed-zero-shot} reports the full zero-shot benchmark table for Qwen2.5-1.5B-Instruct and Qwen2.5-3B-Instruct \cite{qwen2024qwen25}, Llama-3.2-3B-Instruct \cite{meta2024llama32}, and Qwen3-8B \cite{yang2025qwen3}. RULER is reported as the long-context metric \cite{hsieh2024ruler}.

\setlength{\LTleft}{0pt}
\setlength{\LTright}{0pt}
{\widetablefont
\renewcommand{\arraystretch}{0.94}
\setlength{\tabcolsep}{1.0pt}
\begin{longtable}{@{}lll c c c c c c c c c c c c c c c@{}}
\caption{Detailed zero-shot evaluation for all four teacher settings before downstream training. We compare five initialization methods under three retained-layer selection policies and three softmax budgets (25\%, 50\%, and 75\%). Avg is computed over the available short-context metrics in each row; RULER is reported separately as a long-context probe. \textbf{Bold}: best non-teacher PPL (minimum), Avg (maximum), and RULER (maximum) within each model, softmax-budget, and selection block.}
\label{tab:appendix-detailed-zero-shot}\\
\toprule
Softmax \% & Selection & Init & PPL$\downarrow$ & ARC-C & ARC-E & Hella. & PIQA & MMLU & OBQA & RA & WG & BoolQ & LAMB. & COPA & SciQ & Avg & RULER \\
\midrule
\endfirsthead
\multicolumn{18}{@{}l}{\tablename\ \thetable\ (continued)}\\
\toprule
Softmax \% & Selection & Init & PPL$\downarrow$ & ARC-C & ARC-E & Hella. & PIQA & MMLU & OBQA & RA & WG & BoolQ & LAMB. & COPA & SciQ & Avg & RULER \\
\midrule
\endhead
\midrule
\multicolumn{18}{r@{}}{Continued on next page}\\
\endfoot
\bottomrule
\endlastfoot
\multicolumn{18}{@{}l}{\textbf{(a) Qwen2.5-1.5B-Instruct}} \\
\addlinespace
\multicolumn{3}{@{}l}{\textit{Teacher}} & 9.66$\downarrow$ & 46.7 & 76.0 & 68.3 & 76.1 & 60.2 & 40.8 & 38.3 & 62.8 & 77.9 & 60.3 & 83.0 & 94.8 & 64.8 & 86.2 \\
\addlinespace
25\% & Uniform & Baseline & 951064$\downarrow$ & 24.5 & 26.6 & 26.0 & 51.0 & 24.1 & 29.6 & 21.4 & 49.8 & 49.4 & 0.0 & 50.0 & 22.0 & 31.2 & \textbf{0.2} \\
 &  & Zero-Gate & 256980$\downarrow$ & 25.2 & 26.2 & 26.3 & 48.4 & 25.0 & 27.4 & 22.6 & 49.3 & 47.1 & 0.0 & 54.0 & 23.9 & 31.3 & \textbf{0.2} \\
 &  & Small-Gate & 253804$\downarrow$ & 24.3 & 25.3 & 26.3 & 48.5 & 25.1 & 27.4 & 21.1 & 50.2 & 47.4 & 0.0 & 57.0 & 23.7 & 31.4 & \textbf{0.2} \\
 &  & Taylor-Only & 256725$\downarrow$ & 25.3 & 25.7 & 26.5 & 48.7 & 24.9 & 27.6 & 21.6 & 50.7 & 48.0 & 0.0 & 59.0 & 24.9 & 31.9 & \textbf{0.2} \\
\rowcolor{blue!8} &  & \methodname & \textbf{210.9}$\downarrow$ & 24.7 & 31.6 & 28.6 & 53.9 & 22.9 & 29.4 & 22.6 & 51.5 & 54.3 & 0.5 & 59.0 & 38.8 & \textbf{34.8} & \textbf{0.2} \\
\addlinespace
 & \arpolicy & Baseline & 2763$\downarrow$ & 23.5 & 29.8 & 27.2 & 50.8 & 23.2 & 26.0 & 23.0 & 50.3 & 46.4 & 0.1 & 52.0 & 28.1 & 31.7 & 0.0 \\
 &  & Zero-Gate & 543.0$\downarrow$ & 23.9 & 33.5 & 28.3 & 54.6 & 22.9 & 27.6 & 22.9 & 52.0 & 47.5 & 1.2 & 60.0 & 38.6 & 34.4 & 0.0 \\
 &  & Small-Gate & 543.3$\downarrow$ & 24.4 & 33.6 & 28.4 & 55.1 & 22.9 & 27.4 & 23.0 & 53.0 & 47.8 & 1.2 & 60.0 & 38.1 & \textbf{34.6} & 0.0 \\
 &  & Taylor-Only & 542.6$\downarrow$ & 24.2 & 33.7 & 28.4 & 55.0 & 22.9 & 27.6 & 23.3 & 51.9 & 47.7 & 1.3 & 59.0 & 38.5 & 34.5 & 0.0 \\
\rowcolor{blue!8} &  & \methodname & \textbf{163.8}$\downarrow$ & 23.8 & 34.0 & 28.5 & 54.6 & 23.1 & 26.2 & 22.5 & 51.2 & 43.2 & 1.0 & 58.0 & 46.8 & 34.4 & \textbf{0.1} \\
\addlinespace
 & GA-S2 & Baseline & 98568$\downarrow$ & 25.3 & 27.2 & 25.6 & 50.7 & 23.8 & 26.6 & 22.3 & 47.8 & 42.8 & 0.0 & 53.0 & 24.2 & 30.8 & 0.1 \\
 &  & Zero-Gate & 1273$\downarrow$ & 24.1 & 30.8 & 28.6 & 54.4 & 24.4 & 27.4 & 22.9 & 48.2 & 49.5 & 0.8 & 58.0 & 41.7 & 34.2 & 0.3 \\
 &  & Small-Gate & 456.6$\downarrow$ & 22.6 & 36.5 & 30.7 & 56.4 & 23.0 & 27.6 & 22.9 & 51.9 & 40.0 & 2.4 & 61.0 & 57.5 & \textbf{36.0} & \textbf{0.5} \\
 &  & Taylor-Only & 526.8$\downarrow$ & 23.9 & 33.3 & 29.1 & 55.1 & 23.1 & 28.0 & 21.5 & 51.1 & 39.7 & 1.5 & 58.0 & 53.2 & 34.8 & \textbf{0.5} \\
\rowcolor{blue!8} &  & \methodname & \textbf{123.7}$\downarrow$ & 24.0 & 33.6 & 28.4 & 55.1 & 23.6 & 24.8 & 25.2 & 49.8 & 40.2 & 1.2 & 58.0 & 47.6 & 34.3 & 0.2 \\
\addlinespace
50\% & Uniform & Baseline & 2892.9$\downarrow$ & 24.3 & 27.6 & 27.2 & 48.4 & 23.6 & 26.8 & 21.9 & 51.1 & 44.2 & 0.0 & 54.0 & 31.6 & 31.7 & \textbf{0.1} \\
 &  & Zero-Gate & 2377.8$\downarrow$ & 24.8 & 29.8 & 26.9 & 51.6 & 23.4 & 28.6 & 22.3 & 52.3 & 43.8 & 0.0 & 55.0 & 33.5 & 32.7 & \textbf{0.1} \\
 &  & Small-Gate & 2375.3$\downarrow$ & 24.6 & 29.7 & 26.7 & 51.5 & 23.5 & 27.6 & 21.6 & 51.9 & 44.4 & 0.0 & 57.0 & 34.0 & 32.7 & \textbf{0.1} \\
 &  & Taylor-Only & 2423.6$\downarrow$ & 23.6 & 29.4 & 26.7 & 51.5 & 23.2 & 28.2 & 22.8 & 52.0 & 43.6 & 0.0 & 57.0 & 33.1 & 32.6 & \textbf{0.1} \\
\rowcolor{blue!8} &  & \methodname & \textbf{75.2}$\downarrow$ & 24.5 & 39.1 & 31.7 & 57.5 & 23.1 & 28.6 & 23.4 & 50.9 & 45.1 & 3.5 & 63.0 & 58.2 & \textbf{37.4} & \textbf{0.1} \\
\addlinespace
 & \arpolicy & Baseline & 133.0$\downarrow$ & 27.0 & 39.7 & 34.9 & 58.4 & 24.1 & 30.2 & 25.2 & 52.1 & 45.8 & 2.9 & 64.0 & 56.3 & 38.4 & \textbf{0.5} \\
 &  & Zero-Gate & 45.7$\downarrow$ & 29.7 & 48.7 & 45.8 & 68.0 & 24.9 & 35.0 & 27.2 & 52.0 & 48.6 & 14.7 & 73.0 & 72.1 & \textbf{45.0} & \textbf{0.5} \\
 &  & Small-Gate & 45.7$\downarrow$ & 30.0 & 48.6 & 45.7 & 68.5 & 24.8 & 35.8 & 27.2 & 52.3 & 47.9 & 14.7 & 72.0 & 72.0 & \textbf{45.0} & \textbf{0.5} \\
 &  & Taylor-Only & 45.6$\downarrow$ & 29.8 & 48.4 & 45.9 & 68.1 & 24.8 & 35.2 & 27.1 & 52.3 & 47.7 & 14.6 & 72.0 & 72.3 & 44.8 & \textbf{0.5} \\
\rowcolor{blue!8} &  & \methodname & \textbf{36.5}$\downarrow$ & 29.0 & 48.4 & 42.2 & 67.5 & 26.7 & 31.6 & 27.8 & 52.3 & 40.0 & 15.3 & 72.0 & 75.4 & 44.0 & \textbf{0.5} \\
\addlinespace
 & GA-S2 & Baseline & 2698.0$\downarrow$ & 23.0 & 28.5 & 27.3 & 52.1 & 24.1 & 25.6 & 24.8 & 49.3 & 44.2 & 0.4 & 55.0 & 35.3 & 32.5 & 0.2 \\
 &  & Zero-Gate & 100.5$\downarrow$ & 24.2 & 45.5 & 34.0 & 62.8 & 24.3 & 28.0 & 24.5 & 51.2 & 51.9 & 6.2 & 58.0 & 74.1 & 40.4 & 0.6 \\
 &  & Small-Gate & 139.4$\downarrow$ & 27.8 & 45.5 & 35.1 & 65.7 & 24.1 & 29.6 & 27.0 & 51.7 & 46.2 & 9.5 & 68.0 & 68.5 & \textbf{41.6} & \textbf{0.9} \\
 &  & Taylor-Only & 100.5$\downarrow$ & 24.1 & 45.5 & 34.0 & 62.7 & 24.2 & 27.6 & 24.4 & 50.4 & 51.8 & 6.0 & 58.0 & 74.7 & 40.3 & 0.6 \\
\rowcolor{blue!8} &  & \methodname & \textbf{45.5}$\downarrow$ & 22.5 & 42.1 & 31.9 & 58.0 & 24.0 & 27.4 & 25.9 & 52.6 & 44.9 & 12.4 & 61.0 & 75.4 & 39.8 & \textbf{0.9} \\
\addlinespace
75\% & Uniform & Baseline & 43.9$\downarrow$ & 34.4 & 55.1 & 49.5 & 69.0 & 29.3 & 34.2 & 30.5 & 56.4 & 42.8 & 27.3 & 79.0 & 76.2 & 48.6 & 2.7 \\
 &  & Zero-Gate & 30.8$\downarrow$ & 39.3 & 62.9 & 55.1 & 71.8 & 38.8 & 38.0 & 31.7 & 56.2 & 47.3 & 40.5 & 83.0 & 82.0 & 53.9 & 10.1 \\
 &  & Small-Gate & 30.8$\downarrow$ & 39.2 & 62.4 & 55.1 & 72.2 & 38.7 & 37.2 & 31.9 & 56.2 & 47.3 & 40.7 & 82.0 & 81.8 & 53.7 & 10.3 \\
 &  & Taylor-Only & 30.8$\downarrow$ & 39.5 & 62.5 & 55.2 & 72.1 & 38.6 & 37.2 & 32.0 & 56.2 & 47.4 & 40.7 & 83.0 & 82.0 & 53.9 & 10.1 \\
\rowcolor{blue!8} &  & \methodname & \textbf{14.3}$\downarrow$ & 42.2 & 66.2 & 59.3 & 72.5 & 42.2 & 38.8 & 34.5 & 57.7 & 48.1 & 46.5 & 80.0 & 86.1 & \textbf{56.2} & \textbf{14.5} \\
\addlinespace
 & \arpolicy & Baseline & 23.1$\downarrow$ & 34.3 & 59.4 & 50.4 & 70.9 & 30.5 & 35.8 & 31.7 & 53.0 & 44.2 & 26.4 & 76.0 & 78.9 & 49.3 & 6.9 \\
 &  & Zero-Gate & 15.0$\downarrow$ & 38.4 & 60.5 & 59.5 & 73.2 & 43.1 & 38.6 & 31.9 & 54.5 & 48.1 & 36.0 & 81.0 & 84.4 & 54.1 & \textbf{19.4} \\
 &  & Small-Gate & 15.0$\downarrow$ & 38.5 & 60.2 & 59.5 & 73.3 & 43.1 & 39.0 & 31.8 & 53.6 & 48.0 & 36.1 & 81.0 & 84.3 & 54.0 & 19.1 \\
 &  & Taylor-Only & 15.0$\downarrow$ & 38.6 & 60.6 & 59.5 & 73.5 & 43.0 & 38.4 & 31.8 & 54.1 & 48.0 & 36.0 & 81.0 & 84.2 & 54.0 & 19.2 \\
\rowcolor{blue!8} &  & \methodname & \textbf{13.2}$\downarrow$ & 40.6 & 68.2 & 59.0 & 74.3 & 44.5 & 39.4 & 32.1 & 53.5 & 53.8 & 42.3 & 81.0 & 87.8 & \textbf{56.4} & 19.2 \\
\addlinespace
 & GA-S2 & Baseline & 32.9$\downarrow$ & 28.9 & 49.8 & 42.6 & 66.5 & 24.8 & 32.4 & 27.8 & 52.0 & 43.8 & 17.8 & 59.0 & 70.1 & 43.0 & 4.2 \\
 &  & Zero-Gate & 20.1$\downarrow$ & 33.4 & 59.9 & 53.8 & 72.4 & 38.4 & 37.0 & 30.9 & 54.1 & 49.9 & 32.8 & 74.0 & 86.9 & 52.0 & 7.7 \\
 &  & Small-Gate & 22.1$\downarrow$ & 36.9 & 62.8 & 55.9 & 72.4 & 37.4 & 38.4 & 29.5 & 54.6 & 62.9 & 30.8 & 76.0 & 85.2 & 53.6 & 2.3 \\
 &  & Taylor-Only & 21.4$\downarrow$ & 36.1 & 60.8 & 54.4 & 71.8 & 36.6 & 37.2 & 31.5 & 53.7 & 50.0 & 29.7 & 79.0 & 88.1 & 52.4 & 2.4 \\
\rowcolor{blue!8} &  & \methodname & \textbf{14.3}$\downarrow$ & 44.0 & 71.6 & 57.7 & 72.1 & 40.6 & 37.6 & 35.4 & 57.6 & 68.4 & 46.8 & 82.0 & 93.3 & \textbf{58.9} & \textbf{39.3} \\
\addlinespace[6pt]
\multicolumn{18}{@{}l}{\textbf{(b) Qwen2.5-3B-Instruct}} \\
\addlinespace
\multicolumn{3}{@{}l}{\textit{Teacher}} & 8.56$\downarrow$ & 48.0 & 72.9 & 75.0 & 78.2 & 66.4 & 41.8 & 40.2 & 69.5 & 80.2 & 65.7 & 85.0 & 94.7 & 67.3 & 91.3 \\
\addlinespace
25\% & Uniform & Baseline & 13323$\downarrow$ & 25.5 & 26.9 & 26.0 & 52.2 & 22.9 & 27.2 & 21.4 & 50.1 & 37.8 & 0.0 & 60.0 & 23.1 & 31.1 & 0.0 \\
 &  & Zero-Gate & 11503$\downarrow$ & 24.6 & 26.8 & 26.2 & 52.6 & 22.9 & 28.0 & 21.3 & 51.9 & 37.8 & 0.0 & 66.0 & 23.5 & \textbf{31.8} & 0.0 \\
 &  & Small-Gate & 11534$\downarrow$ & 24.4 & 26.8 & 26.2 & 52.6 & 22.9 & 27.8 & 21.1 & 52.1 & 37.8 & 0.0 & 66.0 & 23.4 & \textbf{31.8} & 0.0 \\
 &  & Taylor-Only & 11503$\downarrow$ & 24.4 & 26.8 & 26.1 & 52.2 & 22.9 & 28.2 & 21.4 & 51.7 & 37.8 & 0.0 & 65.0 & 23.6 & 31.7 & 0.0 \\
\rowcolor{blue!8} &  & \methodname & \textbf{241.0}$\downarrow$ & 22.0 & 30.4 & 27.1 & 51.9 & 22.9 & 24.4 & 20.5 & 49.3 & 38.7 & 0.1 & 55.0 & 28.8 & 30.9 & \textbf{0.1} \\
\addlinespace
 & \arpolicy & Baseline & 1815$\downarrow$ & 22.1 & 30.8 & 28.5 & 50.9 & 23.6 & 22.8 & 24.7 & 51.5 & 46.1 & 0.8 & 61.0 & 47.6 & 34.2 & 1.3 \\
 &  & Zero-Gate & 333.8$\downarrow$ & 23.3 & 33.5 & 29.4 & 55.6 & 24.6 & 24.2 & 23.4 & 51.8 & 39.1 & 3.2 & 57.0 & 45.9 & 34.3 & 1.4 \\
 &  & Small-Gate & 337.1$\downarrow$ & 23.7 & 33.4 & 29.5 & 55.4 & 24.4 & 24.6 & 23.4 & 50.8 & 39.7 & 3.3 & 56.0 & 44.9 & 34.1 & \textbf{1.5} \\
 &  & Taylor-Only & 334.9$\downarrow$ & 23.0 & 33.6 & 29.4 & 55.6 & 24.6 & 23.6 & 23.6 & 51.9 & 39.1 & 3.3 & 57.0 & 45.1 & 34.2 & 1.4 \\
\rowcolor{blue!8} &  & \methodname & \textbf{112.8}$\downarrow$ & 22.5 & 35.2 & 29.5 & 56.1 & 24.0 & 24.6 & 24.0 & 50.0 & 52.9 & 2.8 & 61.0 & 67.1 & \textbf{37.5} & 0.4 \\
\addlinespace
 & GA-S2 & Baseline & 14342$\downarrow$ & 26.4 & 25.9 & 26.1 & 50.8 & 22.9 & 29.4 & 21.6 & 50.1 & 37.8 & 0.0 & 53.0 & 21.6 & 30.5 & 0.0 \\
 &  & Zero-Gate & 11638$\downarrow$ & 26.2 & 26.3 & 25.7 & 50.8 & 22.9 & 28.4 & 22.2 & 48.1 & 37.8 & 0.0 & 51.0 & 21.9 & 30.1 & 0.0 \\
 &  & Small-Gate & 12138$\downarrow$ & 26.1 & 26.4 & 25.7 & 50.4 & 22.9 & 27.6 & 21.9 & 47.8 & 37.8 & 0.0 & 52.0 & 22.2 & 30.1 & 0.0 \\
 &  & Taylor-Only & 11571$\downarrow$ & 25.9 & 26.3 & 25.7 & 50.3 & 22.9 & 27.8 & 22.0 & 48.5 & 37.8 & 0.0 & 51.0 & 22.0 & 30.0 & 0.0 \\
\rowcolor{blue!8} &  & \methodname & \textbf{115.9}$\downarrow$ & 21.8 & 35.4 & 29.4 & 55.0 & 23.0 & 26.2 & 23.3 & 50.7 & 57.2 & 4.4 & 56.0 & 56.9 & \textbf{36.6} & \textbf{0.2} \\
\addlinespace
50\% & Uniform & Baseline & 8761$\downarrow$ & 25.8 & 27.2 & 26.0 & 52.1 & 22.9 & 24.8 & 20.8 & 49.6 & 37.8 & 0.0 & 64.0 & 24.1 & 31.3 & 0.0 \\
 &  & Zero-Gate & 8811$\downarrow$ & 26.5 & 27.9 & 25.8 & 51.6 & 22.9 & 26.0 & 21.2 & 49.4 & 37.8 & 0.0 & 62.0 & 24.6 & 31.3 & 0.0 \\
 &  & Small-Gate & 8804$\downarrow$ & 26.1 & 28.2 & 25.9 & 51.4 & 22.9 & 26.2 & 21.0 & 50.4 & 37.8 & 0.0 & 62.0 & 24.4 & 31.4 & 0.0 \\
 &  & Taylor-Only & 8807$\downarrow$ & 26.3 & 28.0 & 25.8 & 51.5 & 22.9 & 26.0 & 21.1 & 49.2 & 37.8 & 0.0 & 62.0 & 24.9 & 31.3 & 0.0 \\
\rowcolor{blue!8} &  & \methodname & \textbf{33.3}$\downarrow$ & 24.2 & 37.2 & 32.7 & 57.1 & 23.0 & 24.4 & 25.5 & 48.6 & 46.2 & 7.4 & 58.0 & 51.8 & \textbf{36.3} & \textbf{0.4} \\
\addlinespace
 & \arpolicy & Baseline & 319.6$\downarrow$ & 23.4 & 37.9 & 34.0 & 56.7 & 24.1 & 27.6 & 25.0 & 50.5 & 43.9 & 3.7 & 64.0 & 55.5 & 37.2 & 2.4 \\
 &  & Zero-Gate & 55.9$\downarrow$ & 29.6 & 54.5 & 44.8 & 66.4 & 27.0 & 34.4 & 27.9 & 50.4 & 48.0 & 15.7 & 78.0 & 73.1 & 45.8 & \textbf{2.5} \\
 &  & Small-Gate & 56.1$\downarrow$ & 29.4 & 54.3 & 44.8 & 66.9 & 26.9 & 34.0 & 28.0 & 51.5 & 48.5 & 15.9 & 78.0 & 73.4 & \textbf{46.0} & \textbf{2.5} \\
 &  & Taylor-Only & 55.8$\downarrow$ & 29.6 & 54.3 & 44.9 & 66.6 & 26.9 & 34.2 & 27.3 & 50.7 & 47.7 & 15.6 & 77.0 & 73.1 & 45.7 & 2.4 \\
\rowcolor{blue!8} &  & \methodname & \textbf{34.2}$\downarrow$ & 26.2 & 49.6 & 40.7 & 64.9 & 26.1 & 33.6 & 26.1 & 51.9 & 57.9 & 19.7 & 70.0 & 71.1 & 44.8 & \textbf{2.5} \\
\addlinespace
 & GA-S2 & Baseline & 8950$\downarrow$ & 25.4 & 26.4 & 26.1 & 51.9 & 23.0 & 28.0 & 22.5 & 50.0 & 37.8 & 0.0 & 56.0 & 21.6 & 30.7 & 0.0 \\
 &  & Zero-Gate & 10477$\downarrow$ & 25.9 & 26.5 & 25.6 & 49.9 & 22.9 & 27.6 & 22.6 & 47.9 & 37.8 & 0.0 & 52.0 & 21.8 & 30.0 & 0.0 \\
 &  & Small-Gate & 10290$\downarrow$ & 25.8 & 26.2 & 25.8 & 50.9 & 22.9 & 27.8 & 22.5 & 48.4 & 37.8 & 0.0 & 52.0 & 21.8 & 30.2 & 0.0 \\
 &  & Taylor-Only & 10393$\downarrow$ & 25.8 & 26.4 & 25.5 & 50.1 & 22.9 & 27.4 & 22.4 & 48.2 & 37.8 & 0.0 & 51.0 & 21.8 & 29.9 & 0.0 \\
\rowcolor{blue!8} &  & \methodname & \textbf{53.3}$\downarrow$ & 25.8 & 45.6 & 34.5 & 59.7 & 23.7 & 30.0 & 25.2 & 50.4 & 50.4 & 15.0 & 69.0 & 68.9 & \textbf{41.5} & \textbf{0.9} \\
\addlinespace
75\% & Uniform & Baseline & 19.7$\downarrow$ & 41.8 & 63.8 & 63.1 & 74.8 & 41.0 & 40.6 & 31.3 & 59.4 & 62.5 & 21.8 & 78.0 & 83.1 & 55.1 & 19.9 \\
 &  & Zero-Gate & 12.9$\downarrow$ & 47.8 & 72.9 & 69.6 & 76.6 & 54.5 & 40.8 & 34.7 & 62.4 & 72.7 & 32.0 & 85.0 & 92.0 & 61.8 & 47.4 \\
 &  & Small-Gate & 12.9$\downarrow$ & 48.0 & 73.2 & 69.7 & 76.9 & 54.7 & 40.6 & 34.7 & 63.1 & 72.8 & 32.0 & 86.0 & 92.1 & \textbf{62.0} & 47.2 \\
 &  & Taylor-Only & 12.9$\downarrow$ & 48.0 & 72.9 & 69.5 & 76.7 & 54.7 & 41.0 & 34.8 & 63.0 & 72.9 & 31.9 & 85.0 & 92.0 & 61.9 & 47.2 \\
\rowcolor{blue!8} &  & \methodname & \textbf{12.0}$\downarrow$ & 47.7 & 72.4 & 69.2 & 75.2 & 54.0 & 41.8 & 35.8 & 62.0 & 73.4 & 32.9 & 83.0 & 88.5 & 61.3 & \textbf{55.9} \\
\addlinespace
 & \arpolicy & Baseline & 28.9$\downarrow$ & 39.2 & 61.5 & 55.7 & 71.1 & 40.2 & 37.8 & 34.6 & 55.6 & 63.0 & 40.7 & 77.0 & 87.3 & 55.3 & 38.7 \\
 &  & Zero-Gate & 13.2$\downarrow$ & 43.9 & 71.3 & 65.4 & 74.1 & 50.7 & 40.0 & 35.5 & 62.2 & 76.2 & 56.1 & 81.0 & 91.3 & \textbf{62.3} & 56.2 \\
 &  & Small-Gate & 13.2$\downarrow$ & 44.2 & 71.1 & 65.3 & 74.2 & 50.8 & 40.2 & 35.5 & 62.2 & 76.3 & 56.3 & 80.0 & 91.4 & \textbf{62.3} & 56.1 \\
 &  & Taylor-Only & 13.2$\downarrow$ & 43.9 & 70.8 & 65.3 & 74.4 & 50.9 & 40.4 & 35.4 & 61.8 & 76.4 & 56.0 & 79.0 & 91.1 & 62.1 & 56.2 \\
\rowcolor{blue!8} &  & \methodname & \textbf{12.5}$\downarrow$ & 43.1 & 71.3 & 63.7 & 73.2 & 49.1 & 41.2 & 36.7 & 62.5 & 78.2 & 56.4 & 80.0 & 89.3 & 62.1 & \textbf{63.7} \\
\addlinespace
 & GA-S2 & Baseline & 13728$\downarrow$ & 25.6 & 26.1 & 26.0 & 51.8 & 22.9 & 28.6 & 21.7 & 48.4 & 37.8 & 0.0 & 56.0 & 22.2 & 30.6 & 0.0 \\
 &  & Zero-Gate & 10588$\downarrow$ & 25.7 & 26.4 & 25.7 & 50.5 & 22.9 & 28.4 & 22.7 & 49.3 & 37.8 & 0.0 & 51.0 & 21.7 & 30.2 & 0.0 \\
 &  & Small-Gate & 10674$\downarrow$ & 25.4 & 26.3 & 25.6 & 51.1 & 22.9 & 27.8 & 22.4 & 48.7 & 37.8 & 0.0 & 52.0 & 21.8 & 30.2 & 0.0 \\
 &  & Taylor-Only & 10572$\downarrow$ & 25.9 & 26.1 & 25.7 & 50.8 & 22.9 & 27.8 & 22.9 & 48.4 & 37.8 & 0.0 & 52.0 & 21.9 & 30.2 & 0.0 \\
\rowcolor{blue!8} &  & \methodname & \textbf{12.9}$\downarrow$ & 48.7 & 72.6 & 67.2 & 75.5 & 44.6 & 38.8 & 38.0 & 61.2 & 72.3 & 54.7 & 79.0 & 86.2 & \textbf{61.6} & \textbf{13.8} \\
\addlinespace[6pt]
\multicolumn{18}{@{}l}{\textbf{(c) Llama-3.2-3B-Instruct}} \\
\addlinespace
\multicolumn{3}{@{}l}{\textit{Teacher}} & 11.05$\downarrow$ & 46.0 & 71.4 & 71.7 & 76.7 & 60.6 & 39.4 & 41.6 & 68.8 & 78.6 & 64.4 & 82.0 & 95.4 & 65.6 & 89.6 \\
\addlinespace
25\% & Uniform & Baseline & 15109$\downarrow$ & 26.1 & 24.9 & 26.2 & 51.8 & 24.4 & 28.4 & 21.4 & 50.6 & 44.4 & 0.0 & 55.0 & 21.1 & 31.2 & 0.0 \\
 &  & Zero-Gate & 1010$\downarrow$ & 21.4 & 31.9 & 28.3 & 55.9 & 23.0 & 28.2 & 22.6 & 52.1 & 37.9 & 0.7 & 51.0 & 44.1 & 33.1 & 0.0 \\
 &  & Small-Gate & 1036$\downarrow$ & 21.2 & 32.2 & 28.4 & 56.6 & 23.0 & 27.8 & 22.4 & 52.1 & 37.9 & 0.7 & 52.0 & 44.4 & 33.2 & \textbf{0.1} \\
 &  & Taylor-Only & 1013$\downarrow$ & 21.1 & 32.3 & 28.4 & 55.7 & 23.0 & 27.8 & 22.4 & 51.8 & 37.9 & 0.7 & 52.0 & 44.4 & 33.1 & \textbf{0.1} \\
\rowcolor{blue!8} &  & \methodname & \textbf{385.3}$\downarrow$ & 22.9 & 32.1 & 28.7 & 55.2 & 22.9 & 26.2 & 22.0 & 49.6 & 38.3 & 1.4 & 59.0 & 51.6 & \textbf{34.2} & \textbf{0.1} \\
\addlinespace
 & \arpolicy & Baseline & 16207$\downarrow$ & 24.2 & 25.4 & 25.7 & 51.7 & 23.9 & 26.6 & 21.7 & 50.0 & 42.3 & 0.0 & 63.0 & 19.9 & 31.2 & \textbf{0.1} \\
 &  & Zero-Gate & 696.6$\downarrow$ & 23.9 & 28.8 & 28.3 & 56.7 & 22.9 & 26.8 & 22.2 & 49.7 & 37.9 & 0.8 & 54.0 & 36.1 & 32.3 & \textbf{0.1} \\
 &  & Small-Gate & 693.7$\downarrow$ & 24.3 & 29.0 & 28.2 & 57.1 & 22.9 & 27.0 & 22.0 & 51.1 & 37.8 & 0.8 & 53.0 & 36.4 & 32.5 & \textbf{0.1} \\
 &  & Taylor-Only & 696.8$\downarrow$ & 24.3 & 29.0 & 28.3 & 56.7 & 22.9 & 26.4 & 22.1 & 49.6 & 37.9 & 0.9 & 55.0 & 36.0 & 32.4 & \textbf{0.1} \\
\rowcolor{blue!8} &  & \methodname & \textbf{132.0}$\downarrow$ & 24.3 & 34.0 & 30.9 & 58.8 & 24.5 & 27.2 & 23.4 & 51.1 & 42.4 & 0.7 & 63.0 & 47.5 & \textbf{35.7} & 0.0 \\
\addlinespace
 & GA-S2 & Baseline & 7649$\downarrow$ & 25.0 & 26.0 & 26.7 & 53.6 & 23.5 & 28.4 & 21.8 & 50.9 & 41.3 & 0.0 & 53.0 & 21.6 & 31.0 & \textbf{0.1} \\
 &  & Zero-Gate & 1347$\downarrow$ & 24.0 & 27.8 & 27.4 & 55.0 & 24.0 & 29.4 & 21.9 & 48.3 & 37.8 & 0.2 & 56.0 & 34.7 & 32.2 & 0.0 \\
 &  & Small-Gate & 1355$\downarrow$ & 24.1 & 27.7 & 27.3 & 55.4 & 23.9 & 29.0 & 21.9 & 49.3 & 37.8 & 0.2 & 55.0 & 34.1 & 32.1 & 0.0 \\
 &  & Taylor-Only & 1348$\downarrow$ & 24.0 & 27.8 & 27.4 & 55.5 & 24.1 & 29.2 & 21.9 & 48.1 & 37.8 & 0.2 & 56.0 & 34.8 & 32.2 & 0.0 \\
\rowcolor{blue!8} &  & \methodname & \textbf{146.4}$\downarrow$ & 27.3 & 37.0 & 35.0 & 61.5 & 23.4 & 27.6 & 22.9 & 51.4 & 56.8 & 0.2 & 63.0 & 36.8 & \textbf{36.9} & \textbf{0.1} \\
\addlinespace
50\% & Uniform & Baseline & 8843$\downarrow$ & 24.4 & 26.6 & 25.9 & 51.9 & 23.3 & 27.4 & 22.9 & 47.7 & 41.0 & 0.0 & 64.0 & 25.5 & 31.7 & 0.1 \\
 &  & Zero-Gate & 595.1$\downarrow$ & 21.5 & 38.4 & 33.4 & 60.4 & 23.8 & 27.2 & 24.0 & 50.1 & 42.3 & 3.1 & 66.0 & 52.6 & 36.9 & 0.2 \\
 &  & Small-Gate & 607.4$\downarrow$ & 22.4 & 38.2 & 33.3 & 60.3 & 23.8 & 27.2 & 24.0 & 49.3 & 42.8 & 3.0 & 65.0 & 51.6 & 36.7 & 0.2 \\
 &  & Taylor-Only & 599.1$\downarrow$ & 21.3 & 38.4 & 33.4 & 60.3 & 23.8 & 27.2 & 24.3 & 49.3 & 42.2 & 3.0 & 66.0 & 52.7 & 36.8 & 0.2 \\
\rowcolor{blue!8} &  & \methodname & \textbf{70.4}$\downarrow$ & 25.1 & 47.7 & 40.1 & 64.6 & 25.2 & 30.2 & 27.6 & 52.7 & 56.4 & 10.1 & 71.0 & 68.4 & \textbf{43.3} & \textbf{0.4} \\
\addlinespace
 & \arpolicy & Baseline & 2033$\downarrow$ & 22.7 & 31.3 & 28.7 & 54.4 & 23.5 & 26.6 & 22.8 & 50.8 & 43.9 & 0.0 & 62.0 & 31.2 & 33.2 & 0.1 \\
 &  & Zero-Gate & 111.1$\downarrow$ & 25.8 & 41.6 & 41.7 & 64.0 & 29.0 & 27.6 & 24.5 & 52.3 & 58.4 & 13.6 & 68.0 & 65.7 & 42.7 & 10.1 \\
 &  & Small-Gate & 111.5$\downarrow$ & 26.4 & 41.8 & 41.7 & 63.7 & 28.8 & 28.6 & 24.6 & 51.5 & 59.0 & 13.5 & 68.0 & 65.0 & 42.7 & 10.3 \\
 &  & Taylor-Only & 111.2$\downarrow$ & 25.8 & 41.7 & 41.8 & 64.1 & 29.0 & 27.8 & 24.2 & 51.3 & 58.3 & 13.6 & 67.0 & 65.6 & 42.5 & 10.1 \\
\rowcolor{blue!8} &  & \methodname & \textbf{33.0}$\downarrow$ & 32.8 & 53.5 & 50.8 & 69.0 & 33.4 & 33.6 & 29.7 & 56.6 & 65.0 & 15.7 & 75.0 & 75.2 & \textbf{49.2} & \textbf{14.3} \\
\addlinespace
 & GA-S2 & Baseline & 2181$\downarrow$ & 25.7 & 27.1 & 26.6 & 51.0 & 23.5 & 26.6 & 23.0 & 48.2 & 39.5 & 0.1 & 53.0 & 25.2 & 30.8 & 0.2 \\
 &  & Zero-Gate & 322.4$\downarrow$ & 23.2 & 34.6 & 31.0 & 56.5 & 22.9 & 26.4 & 23.7 & 51.3 & 37.9 & 3.4 & 52.0 & 63.3 & 35.5 & 0.2 \\
 &  & Small-Gate & 322.4$\downarrow$ & 23.4 & 34.9 & 31.0 & 57.6 & 23.0 & 26.8 & 23.9 & 51.2 & 37.9 & 3.5 & 53.0 & 67.6 & 36.2 & 0.2 \\
 &  & Taylor-Only & 322.2$\downarrow$ & 23.1 & 35.2 & 31.0 & 57.7 & 22.9 & 26.2 & 23.6 & 50.9 & 37.9 & 3.4 & 54.0 & 67.5 & 36.1 & 0.2 \\
\rowcolor{blue!8} &  & \methodname & \textbf{57.6}$\downarrow$ & 35.2 & 59.3 & 56.9 & 70.5 & 43.3 & 34.8 & 29.7 & 57.0 & 61.7 & 19.1 & 79.0 & 84.4 & \textbf{52.6} & \textbf{1.2} \\
\addlinespace
75\% & Uniform & Baseline & 3081$\downarrow$ & 23.1 & 30.9 & 28.2 & 52.1 & 23.5 & 27.2 & 23.6 & 52.0 & 48.6 & 0.1 & 57.0 & 40.9 & 33.9 & 0.2 \\
 &  & Zero-Gate & 25.4$\downarrow$ & 39.7 & 66.9 & 62.1 & 73.1 & 41.7 & 38.6 & 35.1 & 62.4 & 68.6 & 44.1 & 75.0 & 92.5 & 58.3 & \textbf{31.0} \\
 &  & Small-Gate & 25.4$\downarrow$ & 40.0 & 66.7 & 62.1 & 72.8 & 41.8 & 39.0 & 35.5 & 62.9 & 68.5 & 44.2 & 77.0 & 92.6 & 58.6 & 30.9 \\
 &  & Taylor-Only & 25.4$\downarrow$ & 39.8 & 66.8 & 61.9 & 73.3 & 41.8 & 38.8 & 35.5 & 62.8 & 68.5 & 44.1 & 76.0 & 92.5 & 58.5 & 30.9 \\
\rowcolor{blue!8} &  & \methodname & \textbf{21.9}$\downarrow$ & 40.4 & 68.3 & 64.7 & 74.3 & 42.7 & 38.2 & 38.3 & 62.6 & 70.8 & 50.7 & 76.0 & 93.3 & \textbf{60.0} & 24.0 \\
\addlinespace
 & \arpolicy & Baseline & 46.3$\downarrow$ & 32.7 & 56.6 & 51.3 & 69.8 & 33.5 & 35.6 & 29.9 & 53.3 & 52.4 & 21.2 & 71.0 & 80.3 & 49.0 & 9.5 \\
 &  & Zero-Gate & 14.8$\downarrow$ & 42.2 & 68.8 & 66.6 & 75.1 & 52.3 & 39.2 & 38.3 & 65.4 & 67.7 & 45.8 & 77.0 & 89.5 & 60.7 & 55.4 \\
 &  & Small-Gate & 14.8$\downarrow$ & 42.2 & 68.7 & 66.5 & 75.1 & 52.3 & 39.8 & 38.1 & 65.0 & 67.4 & 46.0 & 77.0 & 89.4 & 60.6 & 55.5 \\
 &  & Taylor-Only & 14.8$\downarrow$ & 42.1 & 68.9 & 66.5 & 75.1 & 52.3 & 39.8 & 38.2 & 65.0 & 67.6 & 45.9 & 77.0 & 89.5 & 60.7 & 55.4 \\
\rowcolor{blue!8} &  & \methodname & \textbf{13.7}$\downarrow$ & 44.1 & 71.0 & 68.0 & 75.3 & 54.0 & 39.4 & 39.1 & 66.0 & 71.7 & 50.9 & 79.0 & 91.7 & \textbf{62.5} & \textbf{60.5} \\
\addlinespace
 & GA-S2 & Baseline & 41.6$\downarrow$ & 35.3 & 54.8 & 55.0 & 70.8 & 35.2 & 34.8 & 32.4 & 58.6 & 61.2 & 17.7 & 74.0 & 80.5 & 50.9 & 2.9 \\
 &  & Zero-Gate & 158.4$\downarrow$ & 22.9 & 42.0 & 34.4 & 62.6 & 23.7 & 27.4 & 24.7 & 52.6 & 47.1 & 11.7 & 62.0 & 79.8 & 40.9 & 0.5 \\
 &  & Small-Gate & 157.5$\downarrow$ & 23.2 & 42.3 & 34.4 & 63.1 & 23.7 & 27.2 & 24.6 & 52.2 & 47.2 & 11.7 & 63.0 & 79.7 & 41.0 & 0.5 \\
 &  & Taylor-Only & 158.3$\downarrow$ & 22.9 & 42.2 & 34.5 & 63.1 & 23.6 & 27.4 & 24.4 & 52.4 & 47.1 & 11.7 & 62.0 & 80.0 & 40.9 & 0.5 \\
\rowcolor{blue!8} &  & \methodname & \textbf{14.6}$\downarrow$ & 44.8 & 70.2 & 69.4 & 77.0 & 61.7 & 38.0 & 41.5 & 68.2 & 79.7 & 52.5 & 81.0 & 93.6 & \textbf{64.8} & \textbf{66.3} \\
\addlinespace[6pt]
\multicolumn{18}{@{}l}{\textbf{(d) Qwen3-8B}} \\
\addlinespace
\multicolumn{3}{@{}l}{\textit{Teacher}} & 9.73$\downarrow$ & 56.7 & 80.9 & 74.9 & 77.9 & 74.9 & 41.8 & 41.4 & 68.0 & 86.6 & 64.1 & 85.0 & 96.6 & 70.7 & 94.0 \\
\addlinespace
25\% & Uniform & Baseline & 43609$\downarrow$ & 25.1 & 28.4 & 26.6 & 49.2 & 24.9 & 29.8 & 21.1 & 50.6 & 41.9 & 0.0 & 55.0 & 23.9 & 31.4 & \textbf{0.1} \\
 &  & Zero-Gate & 1106$\downarrow$ & 21.8 & 32.0 & 27.6 & 53.9 & 24.2 & 26.2 & 22.1 & 49.5 & 54.9 & 0.3 & 56.0 & 42.1 & 34.2 & \textbf{0.1} \\
 &  & Small-Gate & 1103$\downarrow$ & 21.5 & 31.6 & 27.7 & 53.6 & 24.3 & 26.6 & 22.2 & 49.0 & 55.1 & 0.3 & 55.0 & 42.4 & 34.1 & 0.0 \\
 &  & Taylor-Only & 1105$\downarrow$ & 21.7 & 31.6 & 27.7 & 53.5 & 24.3 & 26.0 & 22.2 & 48.8 & 54.7 & 0.3 & 57.0 & 42.4 & 34.2 & \textbf{0.1} \\
\rowcolor{blue!8} &  & \methodname & \textbf{166.2}$\downarrow$ & 21.2 & 37.5 & 29.3 & 55.7 & 23.0 & 27.0 & 22.8 & 50.7 & 47.3 & 0.3 & 59.0 & 53.3 & \textbf{35.6} & \textbf{0.1} \\
\addlinespace
 & \arpolicy & Baseline & 3015$\downarrow$ & 23.0 & 27.7 & 26.1 & 51.9 & 24.1 & 29.6 & 21.7 & 50.7 & 38.9 & 0.2 & 55.0 & 32.7 & 31.8 & \textbf{0.1} \\
 &  & Zero-Gate & 482.7$\downarrow$ & 22.2 & 34.6 & 28.2 & 53.3 & 24.0 & 25.4 & 22.9 & 50.5 & 37.9 & 2.5 & 53.0 & 49.5 & 33.6 & 0.0 \\
 &  & Small-Gate & 484.6$\downarrow$ & 22.5 & 34.6 & 28.2 & 54.0 & 23.9 & 25.4 & 23.1 & 50.3 & 37.9 & 2.5 & 50.0 & 49.9 & 33.5 & 0.0 \\
 &  & Taylor-Only & 483.0$\downarrow$ & 22.3 & 34.8 & 28.1 & 53.6 & 23.9 & 25.8 & 23.1 & 50.2 & 37.9 & 2.5 & 52.0 & 49.5 & 33.6 & 0.0 \\
\rowcolor{blue!8} &  & \methodname & \textbf{131.9}$\downarrow$ & 21.8 & 35.0 & 28.7 & 55.1 & 23.1 & 24.4 & 23.5 & 51.2 & 38.6 & 4.8 & 62.0 & 58.3 & \textbf{35.5} & \textbf{0.1} \\
\addlinespace
 & GA-S2 & Baseline & 571090$\downarrow$ & 22.8 & 26.1 & 25.9 & 51.5 & 24.3 & 27.6 & 21.5 & 51.4 & 46.1 & 0.0 & 49.0 & 22.7 & 30.7 & 0.1 \\
 &  & Zero-Gate & 16897$\downarrow$ & 24.7 & 28.2 & 25.9 & 53.2 & 23.2 & 24.4 & 23.0 & 49.8 & 38.0 & 0.0 & 57.0 & 30.9 & 31.5 & 0.0 \\
 &  & Small-Gate & 16915$\downarrow$ & 24.1 & 28.2 & 25.9 & 53.3 & 23.3 & 25.2 & 22.3 & 49.6 & 38.2 & 0.0 & 55.0 & 30.8 & 31.3 & 0.0 \\
 &  & Taylor-Only & 16970$\downarrow$ & 24.7 & 28.0 & 26.2 & 53.0 & 23.2 & 24.6 & 23.1 & 50.2 & 38.0 & 0.0 & 58.0 & 30.7 & 31.6 & 0.0 \\
\rowcolor{blue!8} &  & \methodname & \textbf{204.8}$\downarrow$ & 20.4 & 34.6 & 28.3 & 54.2 & 23.1 & 27.4 & 23.4 & 51.1 & 38.1 & 2.7 & 61.0 & 53.3 & \textbf{34.8} & \textbf{0.2} \\
\addlinespace
50\% & Uniform & Baseline & 361.1$\downarrow$ & 27.7 & 33.8 & 35.9 & 56.9 & 23.9 & 29.4 & 25.6 & 50.6 & 45.1 & 1.1 & 63.0 & 41.2 & 36.2 & 0.2 \\
 &  & Zero-Gate & 53.5$\downarrow$ & 32.9 & 49.5 & 47.3 & 68.5 & 23.8 & 30.2 & 28.2 & 52.2 & 55.8 & 8.1 & 77.0 & 67.2 & 45.1 & 0.3 \\
 &  & Small-Gate & 53.5$\downarrow$ & 33.1 & 49.6 & 47.3 & 68.1 & 23.9 & 30.4 & 28.2 & 52.2 & 55.6 & 8.1 & 76.0 & 66.8 & 44.9 & 0.3 \\
 &  & Taylor-Only & 53.5$\downarrow$ & 33.1 & 49.5 & 47.4 & 68.0 & 23.7 & 30.8 & 28.1 & 51.6 & 55.4 & 8.1 & 77.0 & 66.9 & 45.0 & 0.3 \\
\rowcolor{blue!8} &  & \methodname & \textbf{32.9}$\downarrow$ & 35.8 & 57.4 & 48.9 & 71.8 & 24.9 & 33.2 & 28.3 & 54.0 & 62.6 & 13.7 & 80.0 & 77.6 & \textbf{49.0} & \textbf{0.7} \\
\addlinespace
 & \arpolicy & Baseline & 223.4$\downarrow$ & 23.7 & 42.2 & 33.0 & 57.9 & 25.2 & 25.6 & 23.8 & 49.4 & 39.4 & 5.9 & 55.0 & 63.5 & 37.0 & 0.2 \\
 &  & Zero-Gate & 129.2$\downarrow$ & 27.3 & 50.8 & 34.0 & 61.9 & 25.1 & 27.6 & 25.6 & 49.3 & 40.2 & 8.9 & 64.0 & 71.6 & 40.5 & 0.3 \\
 &  & Small-Gate & 128.9$\downarrow$ & 27.0 & 50.8 & 34.0 & 62.0 & 24.8 & 27.6 & 25.6 & 49.6 & 40.2 & 8.9 & 64.0 & 72.3 & 40.6 & \textbf{0.4} \\
 &  & Taylor-Only & 129.4$\downarrow$ & 27.1 & 50.7 & 34.0 & 62.0 & 24.9 & 27.8 & 25.7 & 49.7 & 40.2 & 8.9 & 64.0 & 72.1 & 40.6 & \textbf{0.4} \\
\rowcolor{blue!8} &  & \methodname & \textbf{52.0}$\downarrow$ & 28.2 & 53.8 & 37.5 & 65.7 & 25.4 & 28.8 & 28.1 & 52.1 & 50.0 & 14.7 & 69.0 & 74.6 & \textbf{44.0} & \textbf{0.4} \\
\addlinespace
 & GA-S2 & Baseline & 1046$\downarrow$ & 27.4 & 39.0 & 34.6 & 59.1 & 24.9 & 32.6 & 23.3 & 50.8 & 43.1 & 2.3 & 57.0 & 48.1 & 36.8 & 0.2 \\
 &  & Zero-Gate & 1043$\downarrow$ & 22.2 & 40.3 & 31.1 & 57.2 & 23.0 & 28.6 & 22.3 & 51.1 & 38.1 & 2.0 & 61.0 & 64.4 & 36.8 & 0.1 \\
 &  & Small-Gate & 373.4$\downarrow$ & 20.8 & 37.5 & 30.0 & 57.9 & 24.7 & 27.2 & 22.9 & 49.0 & 38.5 & 2.1 & 57.0 & 61.6 & 35.8 & 0.0 \\
 &  & Taylor-Only & 1045$\downarrow$ & 21.7 & 40.1 & 31.0 & 57.0 & 23.1 & 28.6 & 22.4 & 51.8 & 38.1 & 1.9 & 62.0 & 64.1 & 36.8 & 0.1 \\
\rowcolor{blue!8} &  & \methodname & \textbf{69.6}$\downarrow$ & 23.4 & 47.4 & 34.4 & 63.9 & 25.4 & 29.6 & 25.4 & 52.6 & 38.4 & 11.3 & 67.0 & 75.9 & \textbf{41.2} & \textbf{0.4} \\
\addlinespace
75\% & Uniform & Baseline & 19.2$\downarrow$ & 47.5 & 69.9 & 61.6 & 73.3 & 55.7 & 39.0 & 37.6 & 62.1 & 72.8 & 32.8 & 77.0 & 88.9 & 59.9 & 42.3 \\
 &  & Zero-Gate & 11.8$\downarrow$ & 52.6 & 76.9 & 67.3 & 76.1 & 60.8 & 38.8 & 39.7 & 63.9 & 80.4 & 49.0 & 79.0 & 93.0 & 64.8 & 61.6 \\
 &  & Small-Gate & 11.8$\downarrow$ & 52.8 & 76.8 & 67.2 & 76.3 & 60.9 & 38.8 & 39.8 & 63.7 & 80.4 & 49.1 & 78.0 & 93.2 & 64.7 & 61.4 \\
 &  & Taylor-Only & 11.8$\downarrow$ & 52.5 & 76.8 & 67.0 & 76.1 & 60.8 & 38.8 & 39.7 & 63.8 & 80.6 & 49.0 & 79.0 & 93.2 & 64.8 & 61.6 \\
\rowcolor{blue!8} &  & \methodname & \textbf{11.4}$\downarrow$ & 54.3 & 77.7 & 67.5 & 75.8 & 61.1 & 38.4 & 39.3 & 65.7 & 81.5 & 49.0 & 82.0 & 93.9 & \textbf{65.5} & \textbf{65.1} \\
\addlinespace
 & \arpolicy & Baseline & 24.2$\downarrow$ & 37.6 & 64.4 & 53.5 & 72.1 & 27.4 & 34.0 & 29.9 & 54.7 & 51.3 & 25.3 & 80.0 & 83.9 & 51.2 & 16.6 \\
 &  & Zero-Gate & 15.9$\downarrow$ & 42.6 & 71.6 & 56.0 & 75.2 & 37.0 & 37.4 & 33.7 & 56.4 & 61.0 & 37.1 & 81.0 & 88.5 & 56.5 & 31.4 \\
 &  & Small-Gate & 15.9$\downarrow$ & 43.4 & 71.6 & 56.0 & 75.2 & 36.9 & 37.8 & 33.6 & 56.4 & 61.3 & 37.1 & 81.0 & 88.4 & 56.6 & 31.2 \\
 &  & Taylor-Only & 15.9$\downarrow$ & 43.1 & 71.7 & 56.0 & 75.2 & 36.7 & 38.0 & 33.2 & 56.1 & 60.8 & 37.0 & 81.0 & 88.6 & 56.5 & 31.1 \\
\rowcolor{blue!8} &  & \methodname & \textbf{13.8}$\downarrow$ & 44.9 & 73.5 & 56.8 & 76.6 & 41.3 & 36.2 & 33.9 & 58.5 & 66.1 & 40.9 & 80.0 & 90.0 & \textbf{58.2} & \textbf{38.2} \\
\addlinespace
 & GA-S2 & Baseline & \textbf{20.6}$\downarrow$ & 39.8 & 64.8 & 58.5 & 72.2 & 43.1 & 34.8 & 33.2 & 57.5 & 62.4 & 35.1 & 78.0 & 88.4 & \textbf{55.6} & \textbf{15.8} \\
 &  & Zero-Gate & 27.2$\downarrow$ & 35.3 & 63.0 & 49.5 & 71.9 & 26.1 & 34.6 & 30.6 & 53.6 & 46.6 & 25.5 & 73.0 & 83.4 & 49.4 & 7.1 \\
 &  & Small-Gate & 27.1$\downarrow$ & 35.4 & 63.0 & 49.6 & 71.5 & 26.2 & 34.2 & 30.6 & 53.7 & 46.6 & 25.6 & 72.0 & 83.4 & 49.3 & 7.0 \\
 &  & Taylor-Only & 27.2$\downarrow$ & 34.9 & 63.1 & 49.6 & 71.9 & 26.3 & 34.4 & 30.5 & 53.3 & 46.5 & 25.4 & 72.0 & 83.6 & 49.3 & 7.0 \\
\rowcolor{blue!8} &  & \methodname & 23.1$\downarrow$ & 36.5 & 62.6 & 48.8 & 71.8 & 29.3 & 37.0 & 30.9 & 53.5 & 53.3 & 24.6 & 75.0 & 76.8 & 50.0 & 6.3 \\
\addlinespace[6pt]
\end{longtable}
}

\FloatBarrier

\paragraph{Recovery dashboards.}
\begin{figure}[H]
\centering
\begin{minipage}[t]{0.48\linewidth}
\centering
\textbf{\large (a) Qwen2.5-1.5B-Instruct}\\[2pt]
{\footnotesize Training dashboard}\\[4pt]
\includegraphics[width=\linewidth]{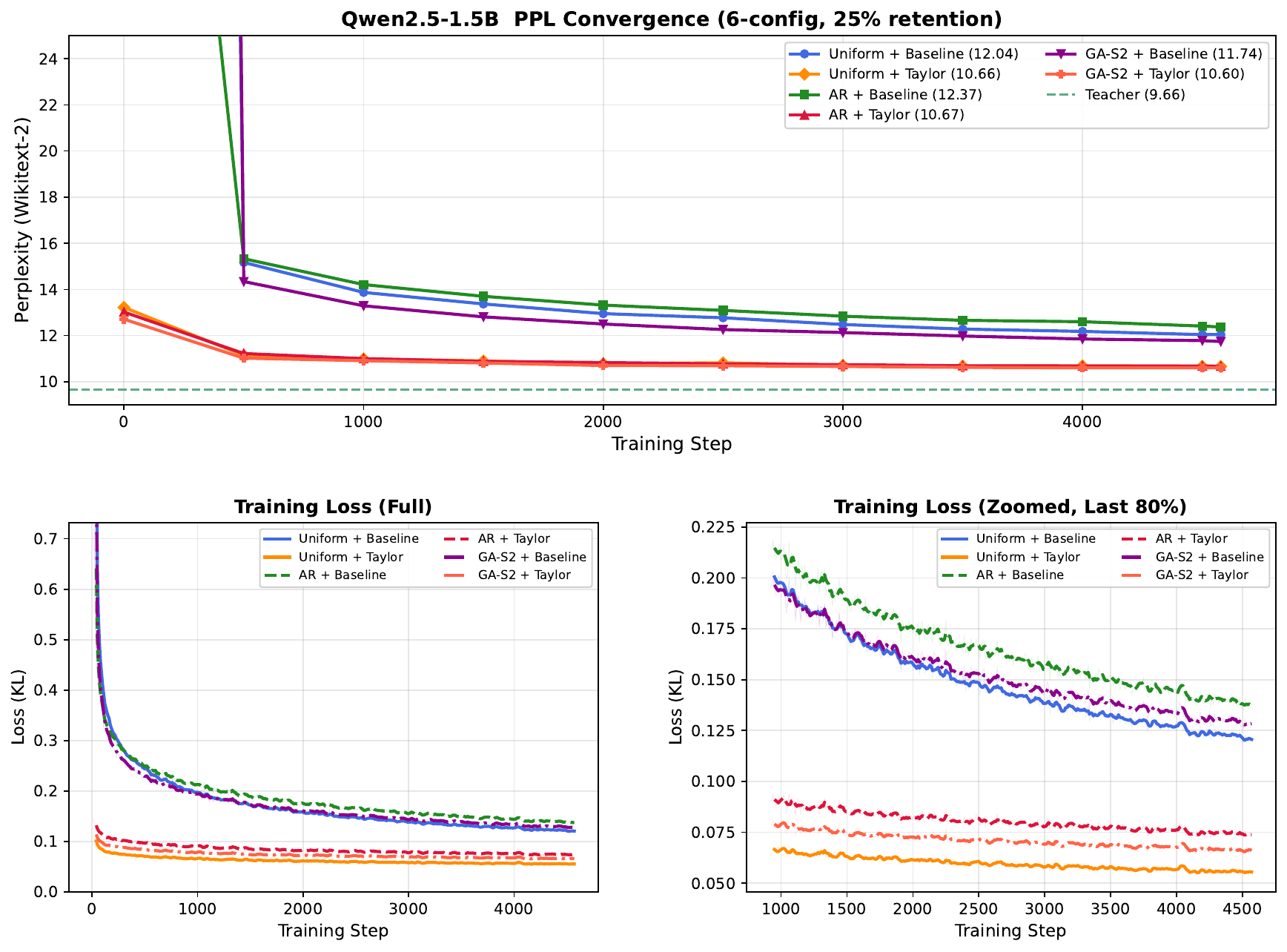}
\end{minipage}
\hfill
\begin{minipage}[t]{0.48\linewidth}
\centering
\textbf{\large (b) Qwen2.5-3B-Instruct}\\[2pt]
{\footnotesize Training dashboard}\\[4pt]
\includegraphics[width=\linewidth]{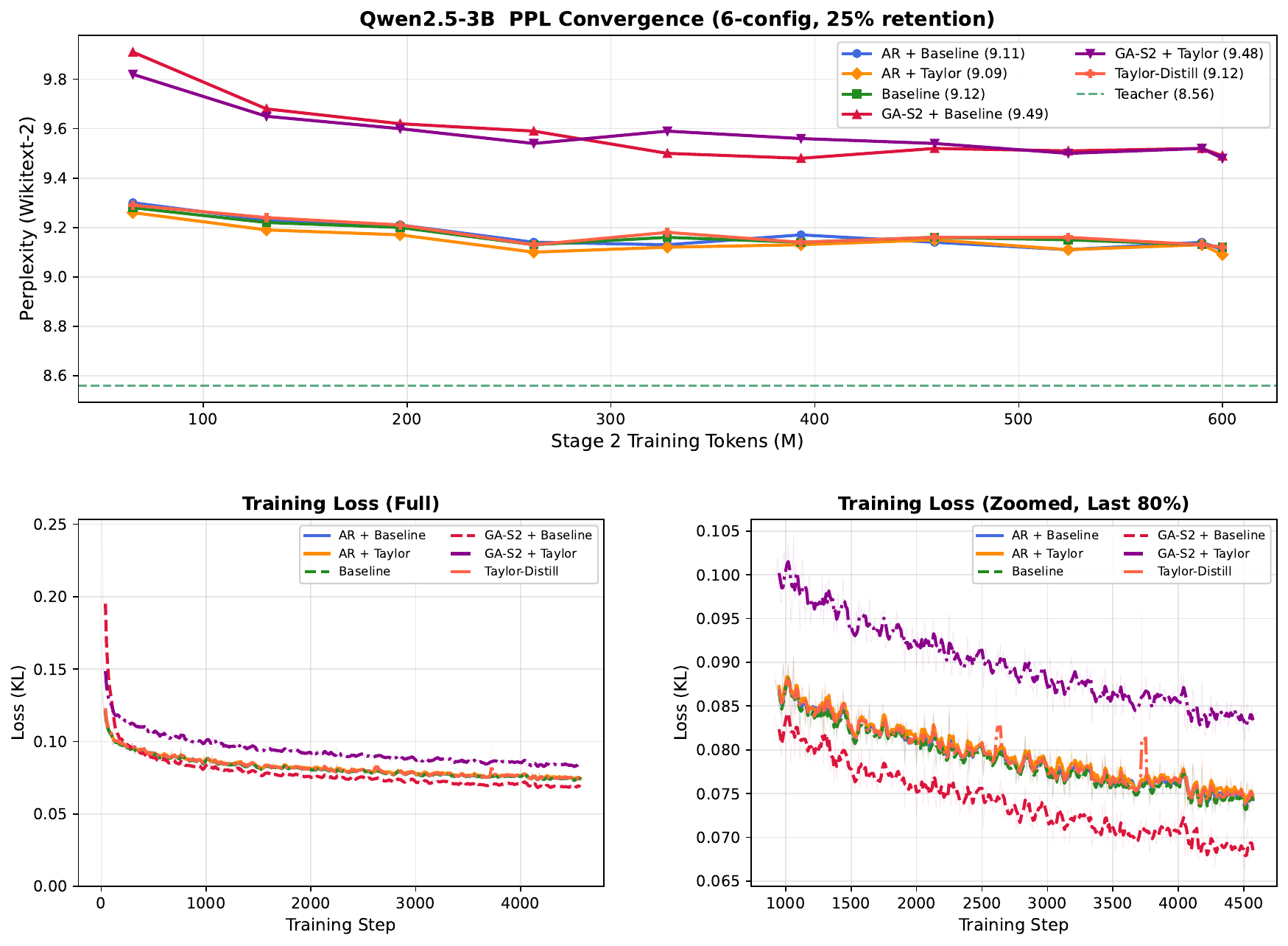}
\end{minipage}
\caption{Recovery dynamics for the two Qwen2.5 teacher settings under matched distillation training.}
\label{fig:qwen25-recovery-curves}
\end{figure}

\FloatBarrier

\paragraph{Recovery details.}
Table~\ref{tab:appendix-detailed-recovery} reports the full checkpoint benchmark table for Qwen2.5-1.5B-Instruct and Qwen2.5-3B-Instruct \cite{qwen2024qwen25}, Llama-3.2-3B-Instruct \cite{meta2024llama32}, and Qwen3-8B \cite{yang2025qwen3}.

\setlength{\LTleft}{0pt}
\setlength{\LTright}{0pt}
{\widetablefont
\renewcommand{\arraystretch}{0.94}
\setlength{\tabcolsep}{1.1pt}
\begin{longtable}{@{}lll c c c c c c c c c c c c c c c@{}}
\caption{Detailed recovery results for all four teacher settings at 100M tokens after Stage 1 and 700M tokens after Stage 2. Avg is shown only when the full short-context set needed for that row is available; RULER is reported separately as a long-context probe. \textbf{Bold}: best non-teacher PPL (minimum), Avg (maximum), and RULER (maximum) within each model block.}
\label{tab:appendix-detailed-recovery}\\
\toprule
Ckpt & Selection & Init & PPL$\downarrow$ & ARC-C & ARC-E & Hella. & PIQA & MMLU & OBQA & RA & WG & BoolQ & LAMB. & COPA & SciQ & Avg & RULER \\
\midrule
\endfirsthead
\multicolumn{18}{@{}l}{\tablename\ \thetable\ (continued)}\\
\toprule
Ckpt & Selection & Init & PPL$\downarrow$ & ARC-C & ARC-E & Hella. & PIQA & MMLU & OBQA & RA & WG & BoolQ & LAMB. & COPA & SciQ & Avg & RULER \\
\midrule
\endhead
\midrule
\multicolumn{18}{r@{}}{Continued on next page}\\
\endfoot
\bottomrule
\endlastfoot
\multicolumn{18}{@{}l}{\textbf{(a) Qwen2.5-1.5B-Instruct}} \\
\addlinespace
\multicolumn{3}{@{}l}{\textit{Teacher}} & 9.66 & 46.7 & 76.0 & 68.3 & 76.1 & 60.2 & 40.8 & 38.3 & 62.8 & 77.9 & 60.3 & 83.0 & 94.8 & 64.8 & 86.2 \\
\addlinespace
\stageoneckpt & Uniform & Baseline & 210.13 & 23.6 & 29.6 & 33.9 & 55.9 & 25.5 & 29.2 & 25.7 & 50.9 & 46.0 & 3.8 & 62.0 & 59.8 & 37.2 & 0.6 \\
 &  & \methodname & 13.54 & 45.9 & 75.9 & 65.9 & 75.4 & 50.8 & 40.8 & 35.4 & 63.1 & 75.8 & 46.7 & 83.0 & 92.8 & 62.6 & 7.7 \\
 & \arpolicy & Baseline & 58.71 & 30.8 & 59.0 & 48.0 & 67.8 & 23.1 & 31.4 & 27.9 & 52.6 & 46.1 & 5.6 & 74.0 & 70.0 & 44.7 & 1.0 \\
 &  & \methodname & 12.89 & 42.7 & 73.4 & 65.4 & 75.3 & 53.4 & 39.8 & 33.4 & 59.1 & 66.5 & 46.8 & 85.0 & 92.5 & 61.1 & 5.0 \\
 & GA-S2 & Baseline & 142.41 & 35.1 & 60.8 & 40.6 & 68.9 & 27.1 & 29.2 & 30.1 & 51.2 & 44.1 & 5.8 & 71.0 & 59.4 & 43.6 & 0.5 \\
 &  & \methodname & 12.54 & 45.6 & 74.5 & 65.5 & 75.8 & 51.9 & 40.8 & 35.0 & 60.5 & 66.8 & 51.6 & 83.0 & 92.5 & 62.0 & 7.1 \\
\addlinespace
\stagetwockpt & Uniform & Baseline & 11.99 & 41.0 & 71.6 & 63.5 & 75.5 & 44.1 & 41.2 & 34.4 & 63.3 & 71.8 & 54.7 & 82.0 & 93.1 & 61.4 & 39.9 \\
 &  & \methodname & 10.76 & 45.6 & 76.8 & 66.6 & 76.2 & 53.1 & 39.8 & 37.4 & 63.9 & 76.5 & 59.0 & 83.0 & 94.6 & \textbf{64.4} & \textbf{59.8} \\
 & \arpolicy & Baseline & 12.23 & 43.4 & 73.6 & 63.3 & 75.4 & 48.5 & 40.4 & 34.8 & 61.3 & 66.9 & 51.8 & 86.0 & 92.0 & 61.4 & 16.8 \\
 &  & \methodname & 10.65 & 45.0 & 75.2 & 66.4 & 75.8 & 55.7 & 39.0 & 35.3 & 62.9 & 72.0 & 56.8 & 85.0 & 94.0 & 63.6 & 38.0 \\
 & GA-S2 & Baseline & 11.96 & 43.9 & 73.9 & 64.1 & 75.5 & 41.6 & 39.0 & 34.8 & 61.6 & 63.5 & 53.7 & 81.0 & 91.1 & 60.3 & 27.0 \\
 &  & \methodname & \textbf{10.57} & 46.8 & 76.3 & 66.4 & 75.9 & 54.6 & 40.0 & 34.9 & 63.0 & 70.9 & 57.5 & 85.0 & 93.7 & 63.8 & 47.6 \\
\addlinespace[6pt]
\multicolumn{18}{@{}l}{\textbf{(b) Qwen2.5-3B-Instruct}} \\
\addlinespace
\multicolumn{3}{@{}l}{\textit{Teacher}} & 8.56 & 48.0 & 72.9 & 75.0 & 78.2 & 66.4 & 41.8 & 40.2 & 69.5 & 80.2 & 65.7 & 85.0 & 94.7 & 67.3 & 91.3 \\
\addlinespace
\stageoneckpt & Uniform & Baseline & 10.78 & 44.3 & 76.0 & 53.6 & 77.7 & 48.5 & 32.2 & 35.7 & 62.5 & 71.6 & 47.2 & 83.0 & 93.5 & 60.5 & 26.3 \\
 &  & \methodname & 10.83 & 44.2 & 75.3 & 53.6 & 77.8 & 51.5 & 33.0 & 34.9 & 62.0 & 72.2 & 47.1 & 84.0 & 93.9 & 60.8 & 29.2 \\
 & \arpolicy & Baseline & 10.32 & 43.8 & 74.7 & 53.3 & 78.0 & 57.5 & 32.2 & 36.5 & 61.6 & 71.3 & 58.6 & 84.0 & 93.4 & 62.1 & 11.5 \\
 &  & \methodname & 10.33 & 44.2 & 75.8 & 53.2 & 78.2 & 57.8 & 32.8 & 35.8 & 61.7 & 69.4 & 58.2 & 85.0 & 93.0 & 62.1 & 11.7 \\
 & GA-S2 & Baseline & 15.97 & 29.7 & 53.7 & 42.7 & 67.9 & 49.8 & 22.6 & 34.5 & 59.0 & 68.8 & 36.7 & 82.0 & 84.3 & 52.6 & 6.6 \\
 &  & \methodname & 11.73 & 42.1 & 73.7 & 53.5 & 78.4 & 57.2 & 32.6 & 35.5 & 60.3 & 62.3 & 54.7 & 83.0 & 92.0 & 60.4 & 4.9 \\
\addlinespace
\stagetwockpt & Uniform & Baseline & 9.09 & 46.9 & 71.9 & 72.8 & 77.8 & 55.1 & 43.6 & 36.5 & 65.8 & 76.8 & 63.2 & 87.0 & 92.5 & 65.8 & 63.8 \\
 &  & \methodname & \textbf{8.95} & 45.4 & 69.9 & 72.8 & 77.6 & 57.0 & 43.2 & 36.9 & 68.0 & 78.0 & 64.1 & 87.0 & 92.5 & 66.0 & 65.7 \\
 & \arpolicy & Baseline & 9.08 & 46.2 & 70.2 & 72.6 & 76.8 & 61.5 & 43.0 & 36.8 & 65.5 & 74.8 & 64.0 & 86.0 & 92.3 & 65.8 & 56.1 \\
 &  & \methodname & 8.98 & 46.9 & 69.3 & 72.9 & 77.1 & 61.8 & 43.2 & 37.8 & 66.4 & 75.3 & 64.3 & 86.0 & 91.7 & \textbf{66.1} & 57.9 \\
 & GA-S2 & Baseline & 9.45 & 46.2 & 71.9 & 72.8 & 77.0 & 55.8 & 42.0 & 36.6 & 67.5 & 77.4 & 63.5 & 86.0 & 92.2 & 65.7 & 59.8 \\
 &  & \methodname & 9.13 & 47.3 & 71.7 & 72.9 & 76.9 & 61.3 & 42.8 & 37.3 & 65.7 & 73.7 & 64.6 & 85.0 & 91.4 & 65.9 & \textbf{70.5} \\
\addlinespace[6pt]
\multicolumn{18}{@{}l}{\textbf{(c) Llama-3.2-3B-Instruct}} \\
\addlinespace
\multicolumn{3}{@{}l}{\textit{Teacher}} & 11.05 & 46.0 & 71.4 & 71.7 & 76.7 & 60.6 & 39.4 & 41.6 & 68.8 & 78.6 & 64.4 & 82.0 & 95.4 & 65.6 & 89.6 \\
\addlinespace
\stageoneckpt & Uniform   & Baseline    & 20.74 & 44.6 & 70.5 & 66.8 & 75.3 & 38.3 & 37.2 & 35.5 & 61.1 & 73.7 & 40.7 & 80.0 & 93.1 & 59.7 & 5.7 \\
 &          & \methodname & 17.41 & 44.8 & 70.4 & 67.5 & 75.5 & 39.9 & 38.2 & 37.4 & 61.3 & 74.7 & 50.6 & 82.0 & 93.0 & 61.3 & 6.6 \\
 & \arpolicy & Baseline   & 20.43 & 47.7 & 73.4 & 66.9 & 76.7 & 31.3 & 39.2 & 34.4 & 59.8 & 72.2 & 28.4 & 82.0 & 90.1 & 58.5 & 4.6 \\
 &          & \methodname & 17.45 & 45.2 & 71.0 & 67.7 & 76.1 & 36.1 & 39.6 & 36.1 & 59.5 & 73.4 & 43.8 & 81.0 & 91.8 & 60.1 & 5.4 \\
 & GA-S2    & Baseline    & 18.87 & 43.7 & 71.0 & 65.8 & 75.5 & 26.5 & 38.8 & 34.0 & 58.1 & 70.1 & 37.9 & 82.0 & 90.7 & 57.8 & 5.0 \\
 &          & \methodname & 16.83 & 45.4 & 71.2 & 67.3 & 76.3 & 31.9 & 38.6 & 34.7 & 63.7 & 73.9 & 39.0 & 80.0 & 92.2 & 59.5 & 4.2 \\
\addlinespace
\stagetwockpt & Uniform  & Baseline     & 12.05 & 46.7 & 72.4 & 69.5 & 76.1 & 49.6 & 37.2 & 41.1 & 65.0 & 77.3 & 61.1 & 81.0 & 94.9 & 64.3 & 60.0 \\
 &          & \methodname & \textbf{11.62} & 45.9 & 72.2 & 69.9 & 76.0 & 49.7 & 39.0 & 41.6 & 66.3 & 77.7 & 61.3 & 83.0 & 94.8 & \textbf{64.8} & \textbf{63.4} \\
 & \arpolicy & Baseline    & 12.22 & 47.3 & 73.2 & 69.6 & 76.5 & 41.4 & 39.6 & 39.0 & 67.2 & 74.5 & 58.6 & 84.0 & 94.3 & 63.8 & 46.4 \\
 &          & \methodname & 11.85 & 44.8 & 71.3 & 69.8 & 76.7 & 45.0 & 40.4 & 40.0 & 67.0 & 75.0 & 60.9 & 82.0 & 95.1 & 64.0 & 47.3 \\
 & GA-S2    & Baseline    & 12.05 & 45.4 & 70.9 & 68.8 & 76.2 & 34.6 & 39.8 & 37.6 & 66.5 & 73.2 & 58.1 & 82.0 & 94.0 & 62.3 & 41.1 \\
 &          & \methodname & 11.92 & 46.2 & 71.7 & 69.5 & 77.1 & 42.1 & 39.4 & 39.0 & 67.1 & 75.7 & 60.2 & 84.0 & 94.9 & 63.9 & 46.5 \\
\addlinespace[6pt]
\multicolumn{18}{@{}l}{\textbf{(d) Qwen3-8B}} \\
\addlinespace
\multicolumn{3}{@{}l}{\textit{Teacher}} & 9.73 & 56.7 & 80.9 & 74.9 & 77.9 & 74.9 & 41.8 & 41.4 & 68.0 & 86.6 & 64.1 & 85.0 & 96.6 & 70.7 & 94.0 \\
\addlinespace
\stageoneckpt & Uniform  & Baseline    &  83.86 & 37.6 & 60.6 & 51.4 & 69.7 & 31.8 & 33.4 & 30.1 & 51.4 & 66.0 &  5.1 & 76.0 & 79.9 & 49.4 & 0.1 \\
 &          & \methodname &  14.92 & 56.1 & 79.9 & 72.8 & 77.3 & 66.2 & 43.6 & 37.1 & 67.9 & 79.0 & 36.9 & 86.0 & 94.7 & 66.5 & 9.2 \\
 & \arpolicy & Baseline   & 172.30 & 32.4 & 51.6 & 49.0 & 68.4 & 27.2 & 32.6 & 29.4 & 51.1 & 57.3 &  3.0 & 69.0 & 65.0 & 44.7 & 0.2 \\
 &          & \methodname &  11.47 & 55.5 & 79.2 & 72.5 & 77.3 & 66.4 & 43.6 & 37.2 & 68.3 & 78.9 & 53.4 & 84.0 & 95.0 & 67.6 & 15.4 \\
 & GA-S2    & Baseline    & 297.36 & 31.1 & 53.4 & 35.9 & 68.3 & 24.2 & 27.4 & 26.0 & 50.0 & 45.7 &  4.8 & 53.0 & 57.4 & 39.8 & 0.1 \\
 &          & \methodname &  11.12 & 56.7 & 79.7 & 72.5 & 77.4 & 63.0 & 41.8 & 36.5 & 68.1 & 78.0 & 53.5 & 87.0 & 95.3 & 67.4 & 11.0 \\
\addlinespace
\stagetwockpt & Uniform  & Baseline    & 10.40 & 56.2 & 80.1 & 73.5 & 78.0 & 67.9 & 42.4 & 36.6 & 69.2 & 82.3 & 61.3 & 89.0 & 96.2 & 69.4 & 60.4 \\
 &          & \methodname & 9.98 & 56.1 & 80.3 & 74.4 & 78.3 & 70.2 & 42.4 & 38.0 & 69.2 & 84.6 & 63.1 & 87.0 & 96.3 & 70.0 & \textbf{76.7} \\
 & \arpolicy & Baseline   & 9.90 & 55.5 & 81.4 & 73.4 & 77.9 & 66.8 & 42.6 & 36.8 & 70.7 & 82.9 & 60.5 & 85.0 & 96.6 & 69.2 & 62.2 \\
 &          & \methodname & \textbf{9.45} & 56.1 & 79.2 & 74.2 & 78.3 & 70.2 & 43.6 & 39.7 & 70.3 & 84.5 & 62.9 & 86.0 & 96.0 & \textbf{70.1} & 69.8 \\
 & GA-S2    & Baseline    & 10.18 & 55.2 & 79.8 & 73.9 & 79.0 & 62.9 & 43.8 & 36.8 & 70.3 & 83.7 & 59.7 & 86.0 & 96.1 & 68.9 & 66.1 \\
 &          & \methodname & 9.46 & 56.9 & 80.7 & 74.0 & 78.3 & 69.0 & 43.0 & 40.1 & 71.4 & 82.9 & 61.9 & 87.0 & 96.0 & \textbf{70.1} & 57.2 \\
\end{longtable}
}

\FloatBarrier

\subsection{Gate Value Statistics}
\label{app:gate_stats}

This section examines whether the gains from \methodname{} are reflected in the internal GDN gate parameters \cite{yang2024gdn}, rather than only in downstream benchmark scores. We analyze the Qwen2.5-3B-Instruct conversion \cite{qwen2024qwen25} with the uniform 25\% full-attention policy. The retained softmax layers are $\{0,4,8,12,16,20,24,28,32\}$, leaving 27 recurrent GDN layers and 432 recurrent heads per initialization variant. We compare four settings: Baseline random GDN initialization, Zero-Gate, Taylor-Only after analytical calibration, and the full \methodname{} initialization after the layer-local alignment step.

\begin{figure}[!htbp]
\centering
\makebox[\linewidth][c]{\includegraphics[width=1.15\linewidth]{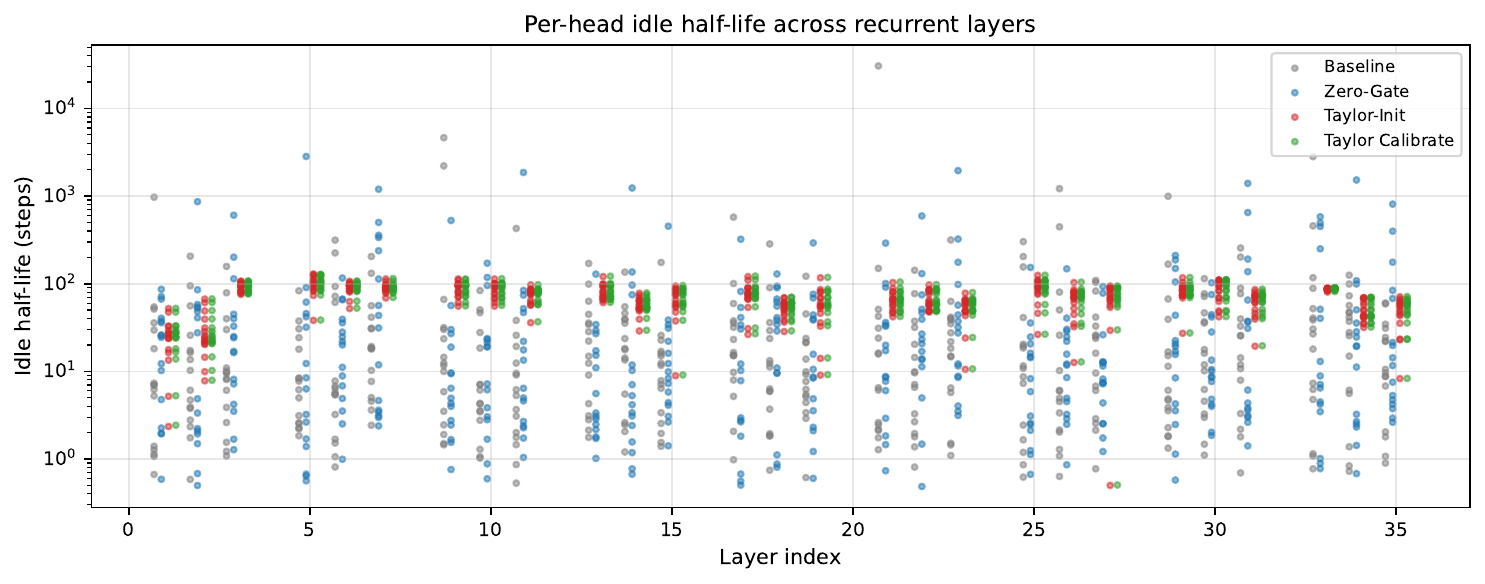}}
\caption{Per-layer idle half-life statistics across recurrent GDN layers. Taylor calibration removes the heavy-tailed random decay outliers and produces a consistent memory timescale across layers.}
\label{fig:gate-halflife}
\end{figure}

\begin{figure}[!htbp]
\centering
\includegraphics[width=\linewidth]{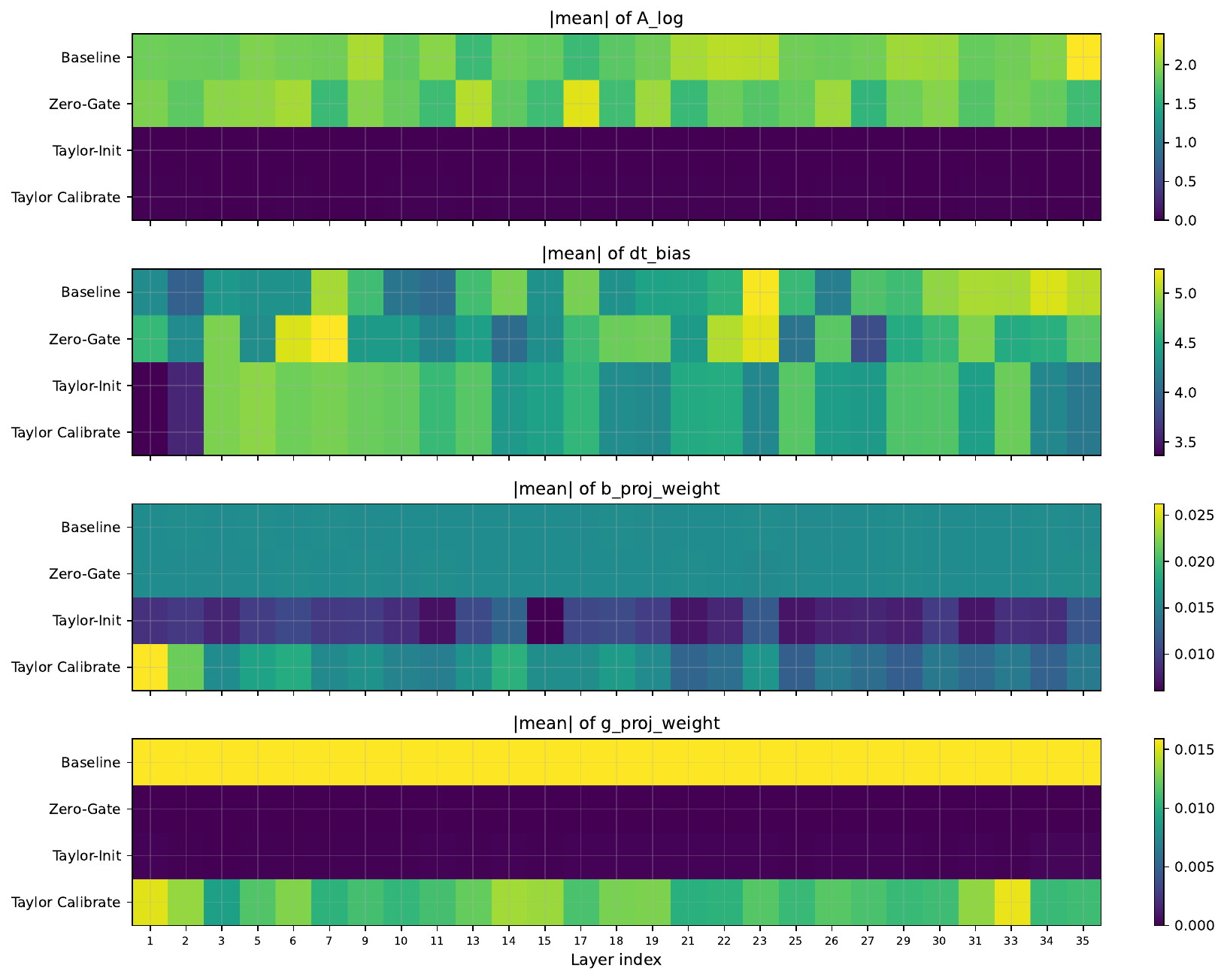}
\caption{Layer-wise magnitudes of the main GDN gate parameters. The Taylor-Only setting suppresses unstable random gate scales, while full \methodname{} restores non-trivial projection-gate magnitude after local alignment.}
\label{fig:gate-summary}
\end{figure}

\begin{figure}[!htbp]
\centering
\makebox[\linewidth][c]{\includegraphics[width=1.35\linewidth]{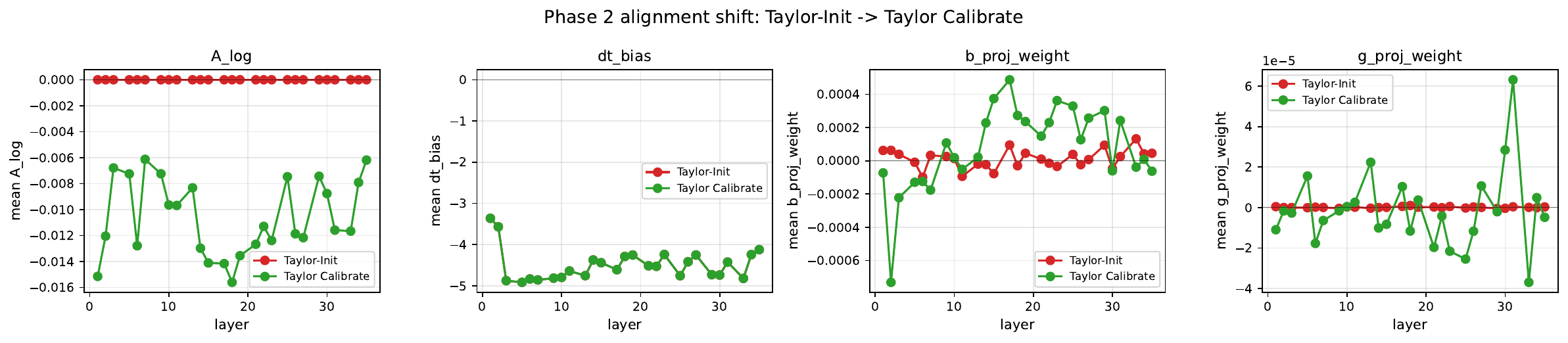}}
\caption{Parameter changes from Taylor-Only to full \methodname{} after layer-local alignment. The recurrent half-life remains nearly fixed, while the write and output projection gates receive most of the alignment-induced change.}
\label{fig:gate-before-after}
\end{figure}

For each recurrent head, we compute the idle half-life implied by the decay parameters,
\begin{equation}
\tau_{1/2}
=
\frac{\ln 2}{\exp(A_{\log})\,\operatorname{softplus}(dt_{\mathrm{bias}})}.
\end{equation}
This statistic measures the memory timescale induced by the initialized recurrence before input-dependent updates are applied. Table~\ref{tab:gate-stat-summary} summarizes the aggregate statistics, while Figures~\ref{fig:gate-halflife}, \ref{fig:gate-summary}, and \ref{fig:gate-before-after} show the corresponding layer-wise distributions and phase-wise changes.

\begin{table}[H]
\centering
\small
\caption{Aggregate gate statistics for the Qwen2.5-3B-Instruct uniform-25\% conversion. Half-life is reported in idle recurrent steps. Gate columns report mean absolute parameter magnitude.}
\label{tab:gate-stat-summary}
\setlength{\tabcolsep}{4.5pt}
\begin{tabular}{@{}lrrrrrrr@{}}
\toprule
Variant & HL mean & HL median & HL p90 & HL max & $|A_{\log}|$ & $|b_{\mathrm{proj}}|$ & $|g_{\mathrm{proj}}|$ \\
\midrule
Baseline & 131.995 & 10.260 & 88.079 & $3.06{\times}10^4$ & 1.923 & 0.016 & 0.016 \\
Zero-Gate & 73.887 & 11.794 & 115.951 & 2834.355 & 1.855 & 0.016 & 0 \\
Taylor-Only & 69.337 & 71.049 & 100.053 & 128.373 & 0 & 0.009 & $1.32{\times}10^{-4}$ \\
\methodname{} & 70.030 & 71.975 & 100.420 & 127.503 & 0.011 & 0.016 & 0.012 \\
\bottomrule
\end{tabular}
\end{table}

The main pattern is that Taylor calibration imposes a much more regular recurrent memory prior than the default GDN initialization. Baseline initialization has a heavy-tailed decay spectrum, with extreme half-life outliers up to roughly $3.06{\times}10^4$ steps. In contrast, the Taylor-calibrated variants concentrate around a median half-life of approximately 71 steps and cap the maximum half-life near 128 steps. This supports the intended role of Phase~1: it sets a controlled memory timescale using teacher attention statistics, rather than relying on the default random decay spectrum.

The layer-local alignment step changes a different part of the parameterization. The mean half-life changes only from 69.337 to 70.030 steps after alignment, a shift of 0.692 steps, indicating that Phase~2 largely preserves the analytically calibrated timescale. Its main effect is instead on the projection gates. The mean absolute magnitude of $g_{\mathrm{proj}}$ increases from $1.32{\times}10^{-4}$ after Taylor-Only to 0.012 after full \methodname{}, an 89.8$\times$ increase, while $b_{\mathrm{proj}}$ increases by 1.73$\times$. Thus Phase~1 places the recurrent state in a controlled memory regime, and Phase~2 activates and orients the write/output gates so that the converted layer produces teacher-like outputs.

\FloatBarrier

\end{document}